\pdfoutput=1 
\documentclass[journal]{IEEEtran}
\usepackage[version=3]{mhchem} 
\usepackage{algorithmic,algorithm}
\usepackage{booktabs}
\usepackage{multirow}
\usepackage{multicol}
\usepackage{array}
\usepackage{tabularx}
\usepackage{amssymb}
\usepackage{flushend}
\usepackage{multirow}
\usepackage{bm}
\usepackage{blindtext}
\usepackage{graphicx}
\usepackage{url}
\usepackage{soul,color}
\usepackage{amsmath}

\makeatletter
\def\hlinew#1{%
  \noalign{\ifnum0=`}\fi\hrule \@height #1 \futurelet
   \reserved@a\@xhline}
\makeatother

\ifCLASSINFOpdf
  
\else

\fi

\begin{document}
%
\title{Deep Learning and Its Applications to Machine Health Monitoring: A Survey}
%
%
%
\author{Rui~Zhao,~Ruqiang~Yan,~Zhenghua~Chen,~Kezhi~Mao,~Peng~Wang,~and~Robert~X.~Gao%
\thanks{
R. Yan is the corresponding author. E-mail: ruqiang@seu.edu.cn \protect\\
This manuscript has been submitted to IEEE Transactions on Neural Networks and Learning Systems}}

%
%

\markboth{Journal of \LaTeX\ Class Files,~Vol.~14, No.~8, August~2015}%
{Shell \MakeLowercase{\textit{et al.}}: Bare Demo of IEEEtran.cls for IEEE Journals}
%



\maketitle

\begin{abstract}
Since 2006, deep learning (DL) has become a rapidly growing research direction, redefining state-of-the-art performances in a wide range of areas such as object recognition, image segmentation, speech recognition and machine translation. In modern manufacturing systems, data-driven
machine health monitoring is gaining in popularity due to the widespread deployment of low-cost sensors and their connection to the Internet. Meanwhile, deep learning provides useful tools for processing and analyzing these big machinery data. The main purpose of this paper is to review and summarize the emerging research work of deep learning on machine health monitoring. After the brief introduction of deep learning techniques, the applications of deep learning in machine health monitoring systems are reviewed mainly from the following aspects: Auto-encoder (AE) and its variants, Restricted Boltzmann Machines and its variants including Deep Belief Network (DBN) and Deep Boltzmann Machines (DBM), Convolutional Neural Networks (CNN) and Recurrent Neural Networks (RNN). Finally, some new trends of
DL-based machine health monitoring methods are discussed.  
\end{abstract}

\begin{IEEEkeywords}
Deep learning, machine health monitoring, big data
\end{IEEEkeywords}

%
\IEEEpeerreviewmaketitle

\section{Introduction}
\IEEEPARstart{I}{ndustrial} Internet of Things (IoT) and data-driven techniques have been revolutionizing manufacturing by enabling  computer networks to gather the huge amount of data from connected machines and turn the big machinery data into actionable information \cite{6748057,jeschkeindustrial,lund2014worldwide}. As a key component in modern manufacturing system, machine health monitoring has fully embraced the big data revolution. Compared to top-down modeling provided by the traditional physics-based models \cite{li2000stochastic, oppenheimer2002physically,yu2014model} , data-driven machine health monitoring systems offer a new paradigm of bottom-up solution for detection of faults after the occurrence of certain failures (diagnosis) and predictions of the future working conditions and the remaining useful life (prognosis) \cite{6748057,jardine2006review}. As we know, the complexity and noisy working condition hinder the construction of physical models. And most of these physics-based models are unable to be updated with on-line measured data, which limits their effectiveness and flexibility. On the other hand, with significant development of sensors, sensor networks and computing systems, data-driven machine health monitoring models have become more and more attractive. To extract useful knowledge and make appropriate decisions from big data, machine learning techniques have been regarded as a powerful solution. As the hottest subfield of machine learning, deep learning is able to act as a bridge connecting big machinery data and intelligent machine health monitoring.

\begin{figure*}
  \centering
  \includegraphics[width=1\textwidth]{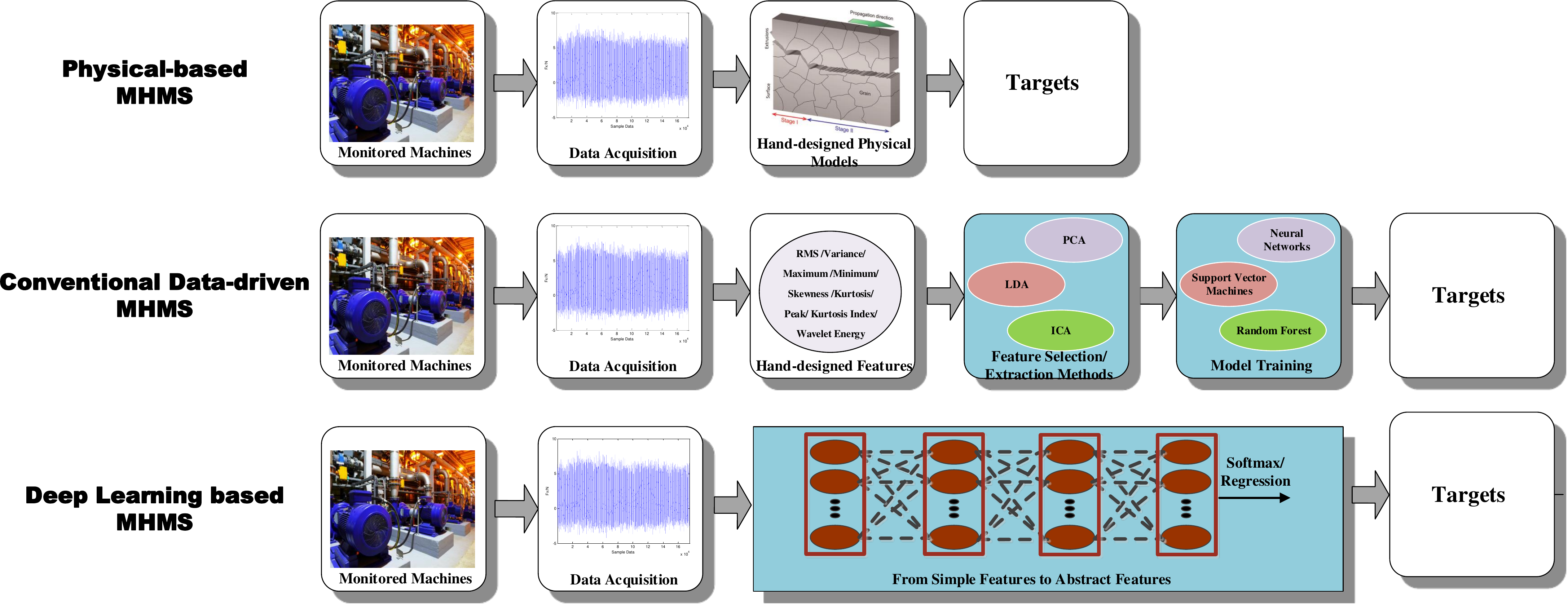}
  \caption{Frameworks showing three different MHMS including Physical Model, Conventional Data-driven Model and Deep Learning Models. Shaded boxes denote data-driven components.}
\label{overview_mhms}
\end{figure*}

As a branch of machine learning, deep learning attempts to model high level representations behind data and classify(predict) patterns via stacking multiple layers of information processing modules in hierarchical architectures. Recently, deep learning has been successfully adopted in various areas such as computer vision, automatic speech recognition, natural language processing, audio recognition and bioinformatics \cite{ren2015faster,collobert2008unified,hinton2012deep,leung2014deep}. In fact, deep learning is not a new idea, which even dates back to the 1940s \cite{888,lecun2015deep}. The popularity of deep learning today can be contributed to the following points:
\begin{enumerate}
\item[*] \textbf{Increasing Computing Power}: the advent of graphics processor unit (GPU), the lowered cost of hardware, the better software infrastructure and the faster network connectivity all reduce the required running time of deep learning algorithms significantly. For example, as reported in \cite{raina2009large}, the time required to learn a four-layer DBN with 100 million free parameters can be reduced from several weeks to around a single day. 
\item[*] \textbf{Increasing Data Size}: there is no doubt that the era of Big Data is coming. Our activities are almost all digitized, recorded by computers and sensors, connected to Internet, and stored in cloud. As pointed out in \cite{6748057} that in industry-related applications such as industrial informatics and electronics, almost 1000 exabytes are generated per year and a 20-fold increase can be expected in the next ten years. The study in \cite{lund2014worldwide} predicts that 30 billion devices will be connected by 2020. Therefore, the huge amount of data is able to offset the complexity increase behind deep learning and improve its generalization capability.   
\item[*] \textbf{Advanced Deep Learning Research}: the first breakthrough of deep learning is the pre-training method in an unsupervised way \cite{hinton2007learning}, where Hinton proposed to pre-train one layer at a time via restricted Boltzmann machine (RBM) and then fine-tune using backpropagation. This has been proven to be effective to train multi-layer neural networks.
\end{enumerate}
Considering the capability of deep learning to address large-scale data and learn high-level representation, deep learning can be a powerful and effective solution for machine health monitoring systems (MHMS). Conventional data-driven MHMS usually consists of the following key parts: hand-crafted feature design, feature extraction/selection and model training. The right set of features are designed, and then provided to some shallow machine learning algorithms including Support Vector Machines (SVM), Naive Bayes (NB), logistic regression \cite{widodo2007support,yan2005degradation,muralidharan2012comparative}. It is shown that the representation defines the upper-bound performances of machine learning algorithms \cite{bengio2013representation}. However, it is difficult to know and determine what kind of good features should be designed. To alleviate this issue, feature extraction/selection methods, which can be regarded as a kind of information fusion, are performed between hand-crafted feature design and classification/regression models \cite{malhi2004pca,wang2016multisensory,7517325}. However, manually designing features for a complex domain requires a great deal of
human labor and can not be updated on-line. At the same time, feature extraction/selection is another tricky problem, which involves prior selection of hyperparameters such as latent dimension. At last, the above three modules including feature design, feature extraction/selection and model training can not be jointly optimized which may hinder the final performance of the whole system. Deep learning based MHMS (DL-based MHMS) aim to extract hierarchical representations from input data by building deep neural networks with multiple layers of non-linear transformations. Intuitively, one layer operation can be regarded as a transformation from input values to output values. Therefore, the application of one layer can learn a new representation of the input data and then, the stacking structure of multiple layers can enable MHMS to learn complex concepts out of simpler concepts that can be constructed from raw input. In addition, DL-based MHMS achieve an end-to-end system, which can automatically learn internal representations from raw input and predict targets. Compared to conventional data driven MHMS, DL-based MHMS do not require extensive human labor and knowledge for hand-crafted feature design. All model parameters including feature module and pattern classification/regression module can be trained jointly. Therefore, DL-based models can be applied to addressing machine health monitoring in a very general way. For example, it is possible that the model trained for fault diagnosis problem can be used for prognosis by only replacing the top softmax layer with a linear regression layer. The comparison between conventional data-driven MHMS and DL-based MHMS is given in Table \ref{compare_ddmhs}. A high-level illustration of the principles behind these three kinds of MHMS discussed above is shown in Figure \ref{overview_mhms}.

Deep learning models have several variants such as Auto-Dncoders \cite{vincent2008extracting}, Deep Belief Network \cite{hinton2006fast}, Deep Boltzmann Machines \cite{salakhutdinov2009deep}, Convolutional Neural Networks \cite{sermanet2012convolutional} and Recurrent Neural Networks \cite{funahashi1993approximation}. During recent years, various researchers have demonstrated success of these deep learning models in the application of machine health monitoring. This paper attempts to provide a wide overview on these latest DL-based MHMS works that impact the state-of-the art technologies. Compared to these frontiers of deep learning including Computer Vision and Natural Language Processing, machine health monitoring community is catching up and has witnessed an emerging research. Therefore, the purpose of this survey article is to present researchers and engineers in the area of machine health monitoring system, a global view of this hot and active topic, and help them to acquire basic knowledge, quickly apply deep learning models and develop novel DL-based MHMS. The remainder of this paper is organized as follows. The basic information on these above deep learning models are given in section \ref{S2}. Then, section \ref{S3} reviews applications of deep learning models on machine health monitoring. Finally, section \ref{S4} gives a brief summary of the recent achievements of DL-based MHMS and discusses some potential trends of deep learning in machine health monitoring.

\begin{table*}
\centering
\caption{Summary on comparison between conventional data-driven MHMS and DL-based MHMS.}
\begin{tabular}{c|c} \hline
\multicolumn{2}{c}{MHMS} \\
\cline{1-2}
  Conventional Data-driven Methods&Deep Learning Methods\\ \hline
 \textit{Expert knowledge and extensive human labor required for
Hand-crafted features}&\textit{End-to-end structure without
hand-crafted features} \\
 \textit{Individual modules are trained step-by-step}&\textit{All parameters are trained jointly} \\
 \textit{Unable to model large-scale data}&\textit{Suitable for large-scale data} \\
   \hline

\end{tabular}
\label{compare_ddmhs}
\end{table*}

\section{Deep Learning}\label{S2}
Originated from artificial neural network,  deep learning is a branch of machine learning which is featured by multiple non-linear processing layers. Deep learning aims to learn hierarchy representations of data. Up to date, there are various deep learning architectures and this research topic is fast-growing, in which new models are being developed even per week. And the community is quite open and there are a number of deep learning tutorials and books of good-quality \cite{Deng2014book,goodfellow2016deep}. Therefore, only a brief introduction to some major deep learning techniques that have been applied in machine health monitoring is given. In the following, four deep architectures including Auto-encoders, RBM, CNN and RNN and their corresponding variants are reviewed, respectively. 

\subsection{Auto-encoders (AE) and its variants}
As a feed-forward neural network, auto-encoder consists of two phases including encoder and decoder. Encoder takes an input $\mathbf{x}$ and transforms it to a hidden representation $\mathbf{h}$ via a non-linear mapping as follows:
 \begin{equation}\label{ae_encoder}
\mathbf{h}= \varphi(\mathbf{W}\mathbf{x}+\mathbf{b})
\end{equation}
where $\varphi$ is a non-linear activation function. Then, decoder maps the hidden representation back to the original representation in a similar way as follows:
 \begin{equation}\label{ae_decoder}
\mathbf{z}= \varphi(\mathbf{W}^{'}\mathbf{h}+\mathbf{b}^{'})
\end{equation}
Model parameters including $\theta=[\mathbf{W}, \mathbf{b}, \mathbf{W}^{'}, \mathbf{b}^{'}]$ are optimized to minimize the reconstruction error between $\mathbf{z}=f_\theta(\mathbf{x})$ and $\mathbf{x}$. One commonly adopted measure for the average reconstruction error over a collection of $N$ data samples is squared error and  the corresponding optimization problem can be written as follows:
 \begin{equation}\label{ae_target}
\min_{\theta}\frac{1}{N}\sum_i^{N}(\mathbf{x}_i-f_\theta(\mathbf{x}_i))^2
\end{equation}
where $\mathbf{x}_i$ is the $i$-th sample. It is clearly shown that AE can be trained in an unsupervised way. And the hidden representation $\mathbf{h}$ can be regarded as a more abstract and meaningful representation for data sample $\mathbf{x}$. Usually, the hidden size should be set to be larger than the input size in AE, which is verified empirically \cite{bengio2007greedy}.

\noindent{\textbf{Addition of Sparsity}}: To prevent the learned transformation is the identity one and regularize auto-encoders, the sparsity constraint is imposed on the hidden units \cite{ng2011sparse}. The corrsponding optimization function is updated as: 

 \begin{equation}\label{sparseae_target}
\min_{\theta}\frac{1}{N}\sum_i^{N}(\mathbf{x}_i-f_\theta(\mathbf{x}_i))^2+\sum_j^{m}KL(p||p_j)
\end{equation}
where $m$ is the hidden layer size and the second term is the summation of the KL-divergence over the hidden units. The KL-divergence on $j$-th hidden neuron is given as:
 \begin{equation}\label{def_kl}
KL(p||p_j) = plog(\frac{p}{p_j})+(1-p)log(\frac{1-p}{1-p_j})
\end{equation}
where $p$ is the predefined mean activation target and $p_j$ is the average activation of the $j$-th hidden neuron over the whole datasets. Given a small $p$, the addition of sparsity constraint can lead the learned hidden representation to be a sparse representation. Therefore, the variant of AE is named sparse auto-encoder.

\noindent{\textbf{Addition of Denoising}}: Different from conventional AE, denoising AE takes a corrupted version of data as input and is trained to reconstruct/denoise the clean input $\mathbf{x}$ from its corrupted sample $\tilde{\mathbf{x}}$. The most common adopted noise is dropout noise/binary masking noise, which randomly sets a fraction of the input features to be zero \cite{vincent2008extracting}. The variant of AE is denoising auto-encoder (DA), which can learn more robust representation and prevent it from learning the identity transformation.  

\noindent{\textbf{Stacking Structure}}: Several DA can be stacked together to form a deep network and learn high-level representations by feeding the outputs of the $l$-th layer as inputs to the $(l+1)$-th layer \cite{vincent2008extracting}. And the training is done one layer greedily at a time.   

\begin{figure*}
  \centering
  \includegraphics[width=1\textwidth]{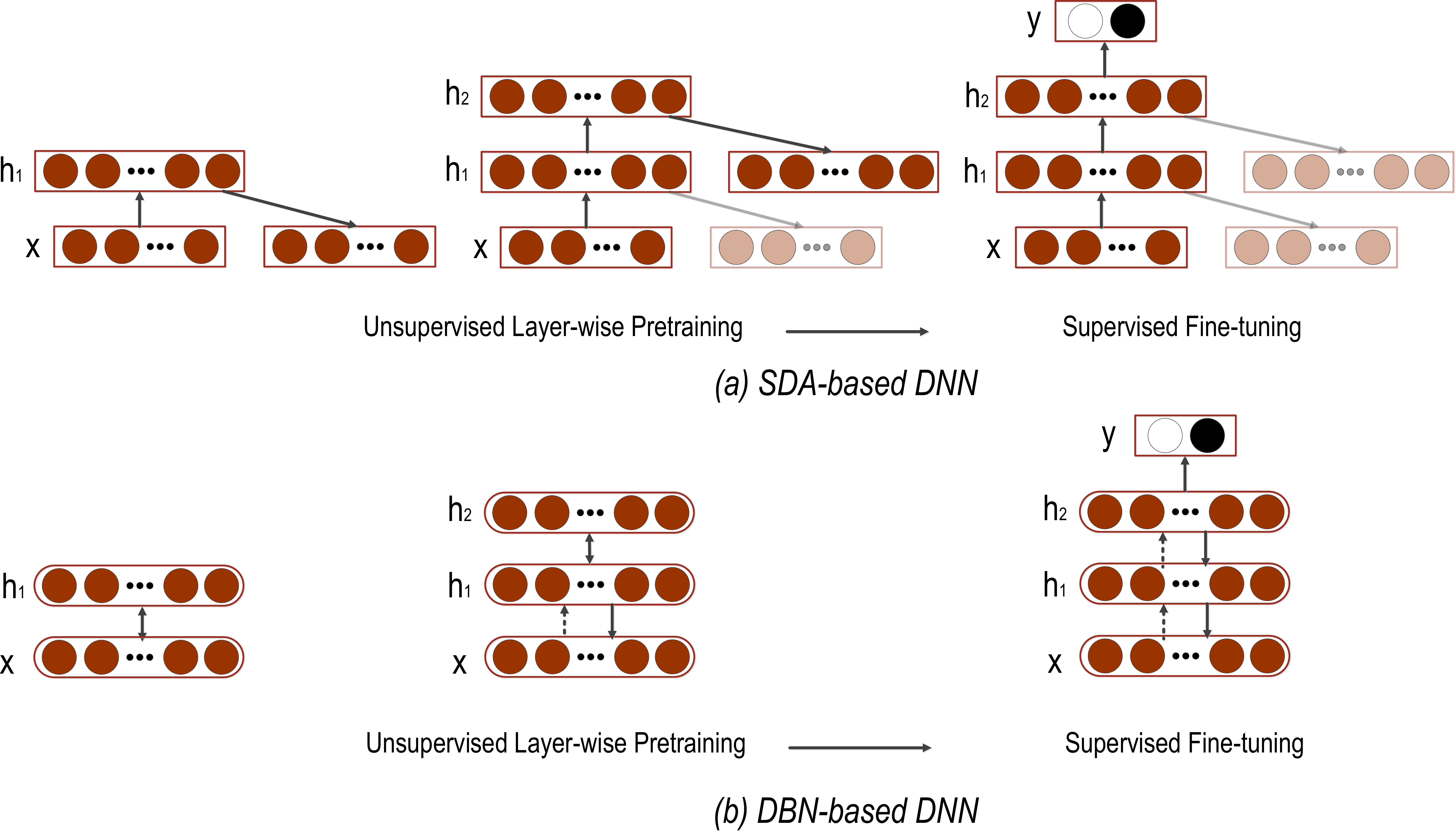}
  \caption{Illustrations for Unsupervised Pre-training and Supervised Fine-tuning of SAE-DNN (a) and DBN-DNN (b).}
\label{overview_ae}
\end{figure*}

Since auto-encoder can be trained in an unsupervised way, auto-encoder, especially stacked denoising auto-encoder (SDA), can provide an effective pre-training solution via initializing the weights of deep neural network (DNN) to train the model. After layer-wise pre-training of SDA, the parameters of auto-encoders can be set to the initialization for all the hidden layers of DNN. And then, the supervised fine-tuning is performed to minimize prediction error on a labeled training data. Usually, a softmax/regression layer is added on top of the network to map the output of the last layer in AE to targets. The whole process is shown in Figure \ref{overview_ae}. The pre-training protocol based on SDA can make DNN models have better convergence capability compared to arbitrary random initialization.  

\subsection{RBM and its variants}
	As a special type of Markov random field, restricted Boltzmann machine (RBM) is a two-layer neural network forming a bipartite graph that consists of two groups of units including visible units $\mathbf{v}$ and hidden units $\mathbf{h}$ under the constrain that there exists a symmetric connection between visible units and hidden units and there are no connections between nodes with a group.

Given the model parameters $\theta=[\mathbf{W},\mathbf{b},\mathbf{a}]$, the energy function can be given as:

\begin{equation}\label{rbm_eng}
E(\mathbf{v},\mathbf{h};\theta) = -\sum_{i=1}^{I}\sum_{j=1}^{J}w_{ij}v_ih_j-\sum_{i=1}^{I}b_iv_i-\sum_{j=1}^{J}a_jh_j
\end{equation}
 that $w_{ij}$ is the connecting weight between visible unit $v_i$, whose total number is $I$ and hidden unit $h_j$ whose total number is $J$, $b_i$ and $a_j$ denote the bias terms for visible units and hidden units, respectively. The joint distribution over all the units is calculated based on the energy function $E(\mathbf{v},\mathbf{h};\theta)$ as:
\begin{equation}\label{rbm}
p(\mathbf{v},\mathbf{h};\theta) = \frac{exp(-E(\mathbf{v},\mathbf{h};\theta))}{Z}
\end{equation}
where $Z=\sum_{\mathbf{h};\mathbf{v}}exp(-E(\mathbf{v},\mathbf{h};\theta))$ is the partition function or normalization factor. Then, the conditional probabilities of hidden and visible units $\mathbf{h}$ and $\mathbf{v}$ can be calculated as:

\begin{equation}\label{rbm_probh}
p(h_j=1|v;\theta) = \delta(\sum_{i=1}^{I}w_{ij}v_i+a_j)
\end{equation}

\begin{equation}\label{rbm_probv}
p(v_i=1|v;\theta) = \delta(\sum_{j=1}^{J}w_{ij}h_j+b_i)
\end{equation}
where $\delta$ is defined as a logistic function i.e., $\delta(x)=\frac{1}{1+exp(x)}$. RBM are trained to maximize the joint probability. The learning of W is done through a method called contrastive divergence (CD).

\begin{figure}
  \centering
  \includegraphics[width=0.5\textwidth]{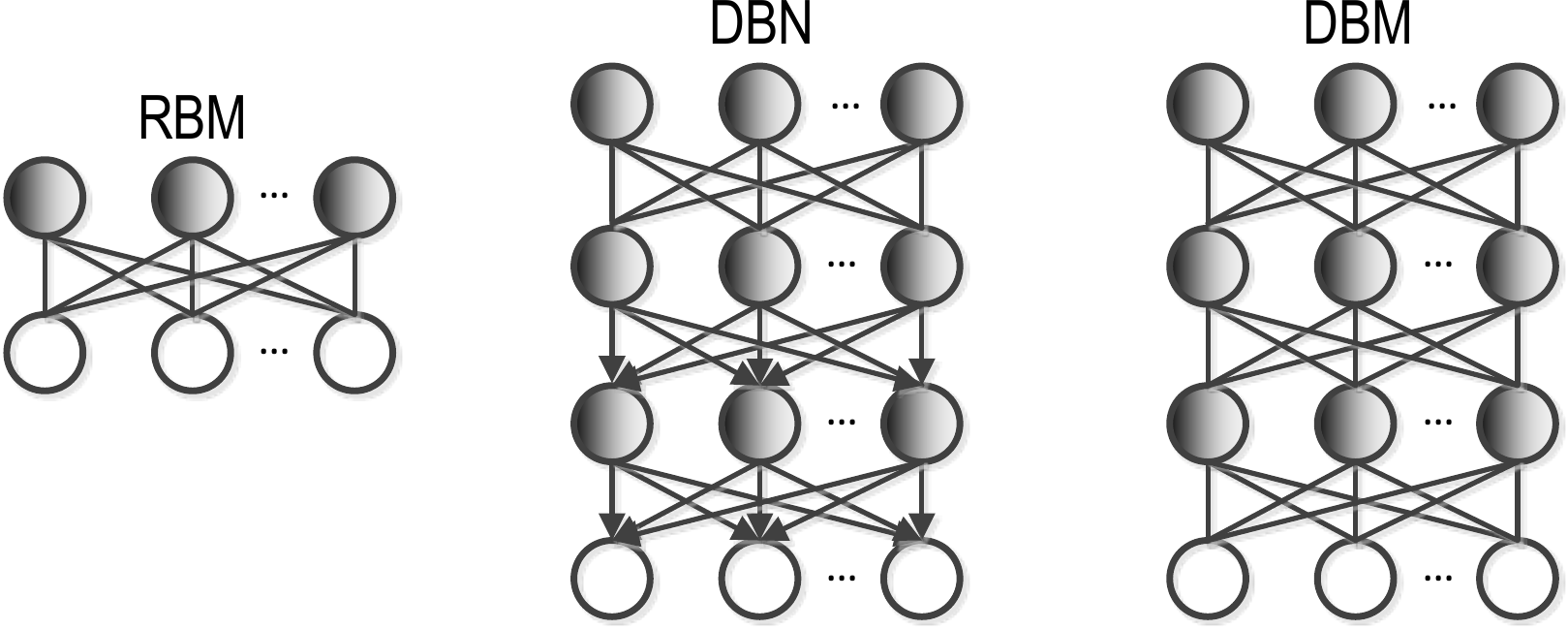}
  \caption{Frameworks showing RBM, DBN and DBM. Shaded boxes denote hidden untis.}
\label{overview_rbm}
\end{figure}

\noindent{\textbf{Deep Belief Network}}: Deep belief networks (DBN) can be constructed by stacking multiple RBMs, where the output of the $l$-th layer (hidden units) is used as the input of the $(l+1)$-th layer (visible units). Similar to SDA, DBN can be trained in a greedy layer-wise unsupervised way. After pre-training, the parameters of this deep architecture can be further fine-tuned with respect to a proxy for the DBN log- likelihood, or with respect to labels of training data by adding a softmax layer as the top layer, which is shown in Figure \ref{overview_ae}.(b).

\noindent{\textbf{Deep Boltzmann Machine}}: Deep Boltzmann machine (DBM) can be regarded as a deep structured RMBs where hidden units are grouped into a hierarchy of layers instead of a single layer. And following  the RMB's connectivity constraint, there is only full connectivity between subsequent layers and no connections within layers or between non-neighbouring layers are allowed. The main difference between DBN and DBM lies that DBM is fully undirected graphical model, while DBN is mixed directed/undirected one. Different from DBN that can be trained layer-wisely, DBM is trained as a joint model. Therefore, the training of DBM is more computationally expensive than that of DBN. 

\subsection{Convolutioanl Neural Network}\label{intro_cnn}
Convolutional neural networks (CNNs) were firstly proposed by LeCun \cite{le1990handwritten} for image processing, which is featured by two key properties: spatially shared weights and spatial pooling. CNN models have shown their success in various computer vision applications \cite{le1990handwritten,jarrett2009best,krizhevsky2012imagenet} where input data are usually 2D data. CNN has also been introduced to address sequential data including Natural Language Processing and Speech Recognition \cite{abdel2012applying,kim2014convolutional}.

CNN aims to learn abstract features by alternating and stacking convolutional kernels and pooling operation. In CNN, the convolutional layers (convolutional kernels) convolve multiple local filters with raw input data and generate invariant local features and the subsequent pooling layers extract most significant features with a fixed-length over sliding windows of the raw input data. Considering 2D-CNN have been illustrated extensively in previous research compared to 1D-CNN, here, only the mathematical details behind 1D-CNN is given as follows: 

Firstly, we assume that the input sequential data is $\mathbf{x} = [ \mathbf{x}_{1}, \dots, \mathbf{x}_{T} ]$ that $T$ is the length of the sequence and $\mathbf{x}_i\in {\mathbb{R}^{d}}$ at each time step.

\noindent{\bf Convolution}: the dot product between a filter vector $\mathbf{u}\in {\mathbb{R}^{md}}$ and an concatenation vector representation $\mathbf{x}_{i:i+m-1}$ defines the convolution operation as follows:
\begin{equation}\label{convoperation}
c_{i}=\varphi(\mathbf{u}^{T}\mathbf{x}_{i:i+m-1}+b)
\end{equation}
where ${\mathbf{*}}^{T}$ represents the transpose of a matrix ${\mathbf{*}}$, $b$ and $\varphi$ denotes bias term and non-linear activation function, respectively. $\mathbf{x}_{i:i+m-1}$ is a $m$-length window starting from the $i$-th time step, which is described as:
\begin{equation}\label{windows}
\mathbf{x}_{i:i+m-1} =\mathbf{x}_i\oplus\mathbf{x}_{i+1}\oplus\dots\oplus\mathbf{x}_{i+m-1}
\end{equation}

As defined in Eq. \eqref{convoperation}, the output scale $c_i$ can be regarded as the activation of the filter $\mathbf{u}$ on the corresponding subsequence $\mathbf{x}_{i:i+m-1}$. By sliding the filtering window from the beginning time step to the ending time step, a feature map as a vector can be given as follows:  
\begin{equation}\label{featuremap}
\mathbf{c}_j=\left[c_1,c_2,\dots,c_{l-m+1}\right]
\end{equation}
where the index $j$ represents the $j$-th filter. It corresponds to multi-windows as $\{\mathbf{x}_{1:m},\mathbf{x}_{2:m+1},\dots,\mathbf{x}_{l-m+1:l}\}$.

\noindent{\bf Max-pooling}: Pooling layer is able to reduce the length of the feature map, which can further minimize the number of model parameters. The hyper-parameter of pooling layer is pooling length denoted as $s$. MAX operation is taking a max over the $s$ consecutive values in feature map $\mathbf{c}_j$.

\begin{figure*}
  \centering
  \includegraphics[width=0.75\textwidth]{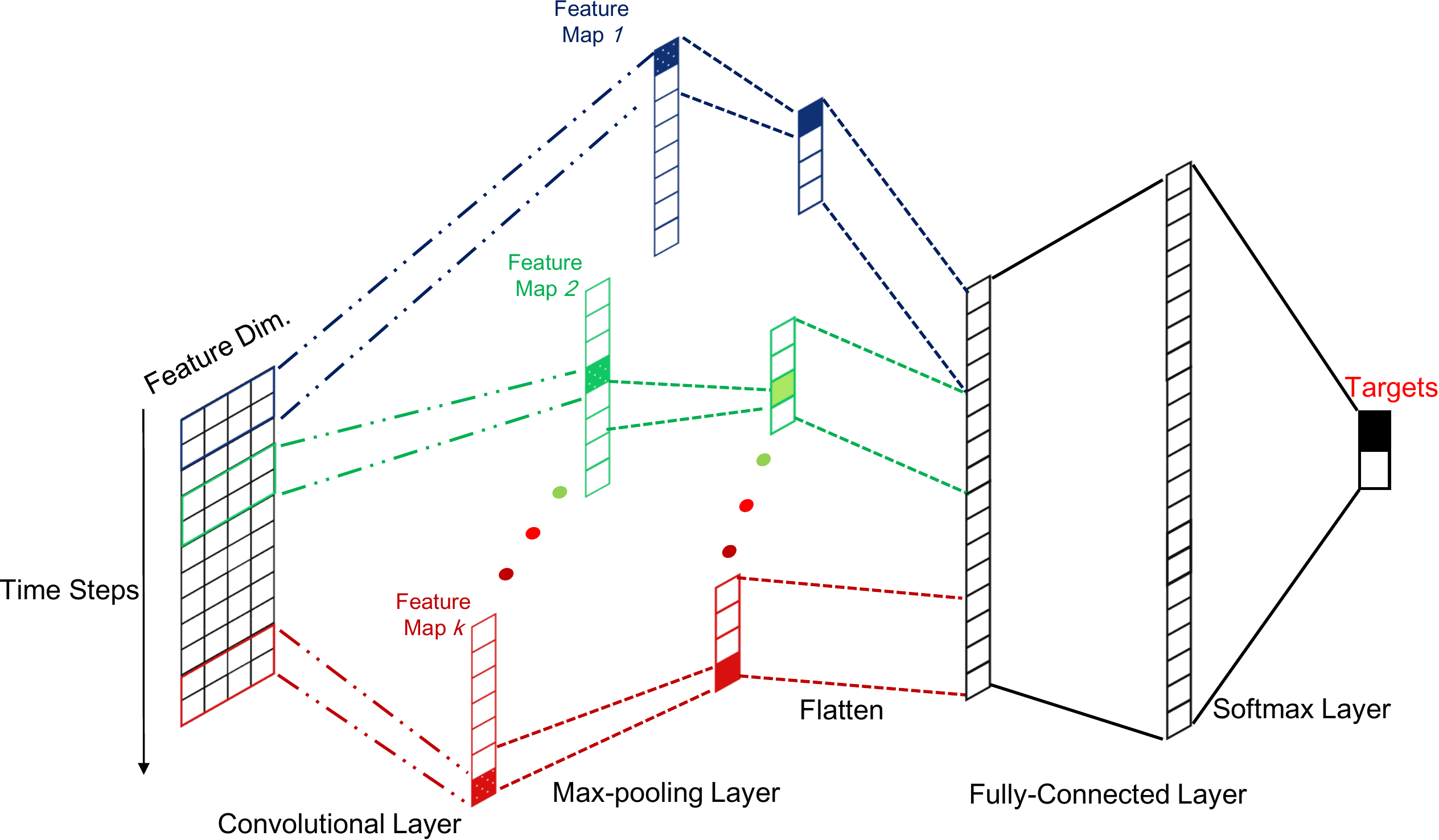}
  \caption{Illustrations for one-layer CNN that contains one convolutional layer, one pooling layer, one fully-connected layer, and one softmax layer.}
\label{overview_cnn}
\end{figure*}

Then, the compressed feature vector can be obtained as: 

\begin{equation}\label{max}
\mathbf{h} = \left[h_1,h_2,\dots,h_{\frac{l-m}{s}+1}\right]
\end{equation}
where $h_j= \max(c_{(j-1)s},c_{(j-1)s+1},\dots,c_{js-1})$. Then, via alternating the above two layers: convolution and max-pooling ones, fully connected layers and a softmax layer are usually added as the top layers to make predictions. To give a clear illustration, the framework for a one-layer CNN has been displayed in Figure. \ref{overview_cnn}.

\subsection{Recurrent Neural Network}
As stated in \cite{888}, recurrent neural networks (RNN) are the deepest of all neural networks, which can generate and address memories of arbitrary-length sequences of input patterns. RNN is able to build connections between units from a directed cycle. Different from basic neural network: multi-layer perceptron that can only map from input data to target vectors, RNN is able to map from the entire history of previous inputs to target vectors in principal and allows a memory of previous inputs to be kept in the network's internal state. RNNs can be trained via backpropagation through time for supervised tasks with sequential input data and target outputs \cite{jaeger2002tutorial,funahashi1993approximation,giles1992learning}. 

\begin{figure}
  \centering
  \includegraphics[width=0.5\textwidth]{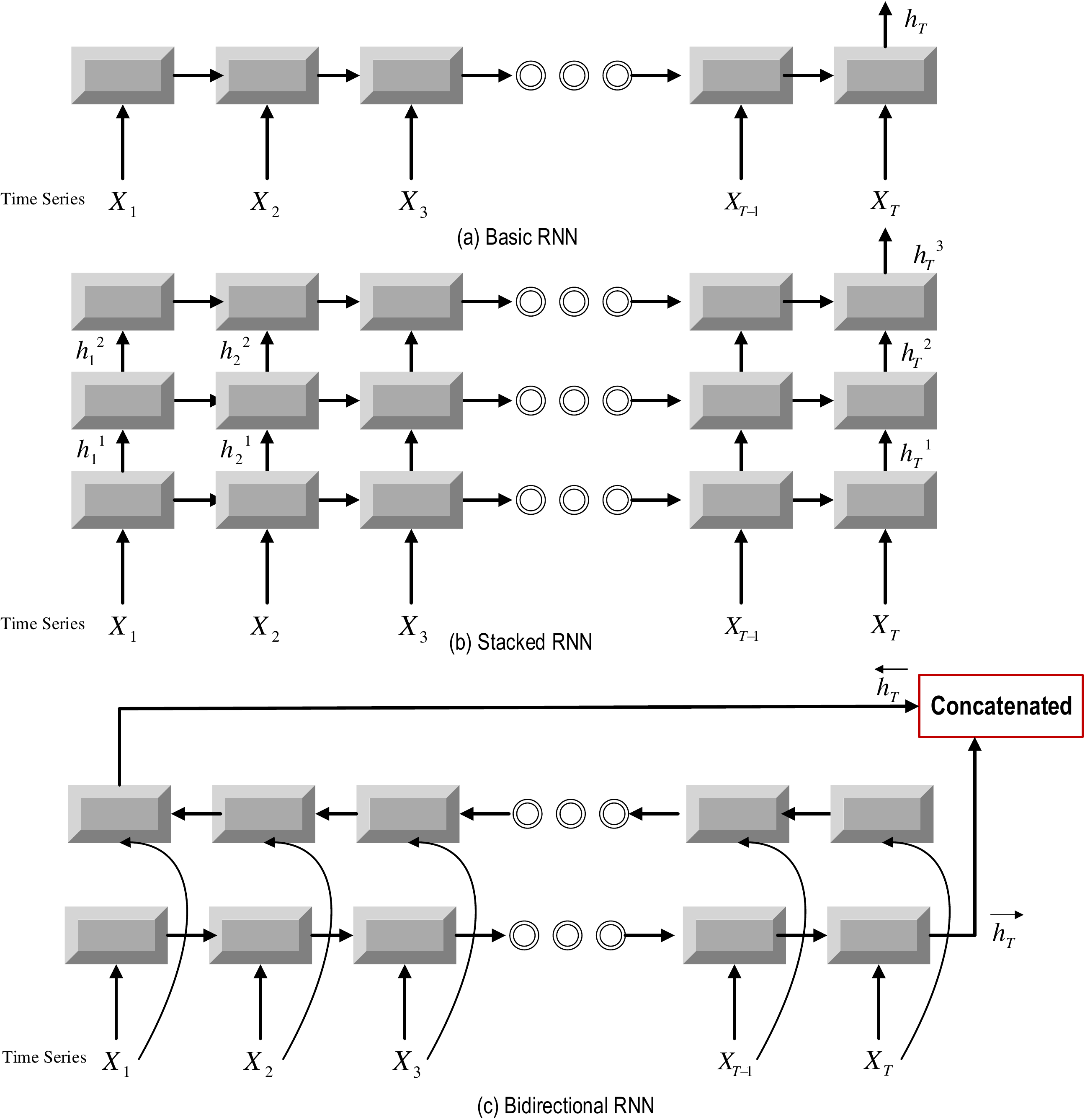}
  \caption{Illustrations of normal RNN, stacked RNN and bidirectional RNN.}
\label{illus_rnnvar}
\end{figure}

RNN can address the sequential data using its internal memory, as shown in Figure \ref{illus_rnnvar}. (a). The transition function defined in each time step $t$ takes the current time information $\mathbf{x}_t$ and the previous hidden output $\mathbf{h}_{t-1}$ and updates the current hidden output as follows:
\begin{equation}\label{RNN_overview}
\mathbf{h}_t{} = \mathbb{H}(\mathbf{x}_{t}, \mathbf{h}_{t-1})
\end{equation} 
where $\mathbb{H}$ defines a nonlinear and differentiable transformation function. After processing the whole sequence, the hidden output at the last time step i.e. $\mathbf{h}_{T}$ is the learned representation of the input sequential data whose length is $T$. A conventional Multilayer perceptron (MLP) is added on top to map the obtained representation $\mathbf{h}_T$ to targets.

Various transition functions can lead to various RNN models. The most simple one is vanilla RNN that is given as follows:
\begin{equation}\label{RNN_nrom}
\mathbf{h}_{t} = \varphi(\mathbf{W}\mathbf{x}_{t}+\mathbf{H}\mathbf{h}_{t-1}+\mathbf{b})
\end{equation} 
where $\mathbf{W}$ and $\mathbf{H}$ denote transformation matrices and $\mathbf{b}$ is the bias vector. And $\varphi$ denote the nonlinear activation function such as sigmoid and tanh functions. Due to the vanishing gradient problem during backpropagation for model training, vanilla RNN may not capture long-term dependencies. Therefore, Long-short term memory (LSTM) and gated recurrent neural networks (GRU) were presented to prevent backpropagated errors from vanishing or exploding \cite{hochreiter1997long,gers2000learning,gers2002learning,cho2014learning,chung2014empirical}. The core idea behind these advanced RNN variants is that gates are introduced to avoid the long-term dependency problem and enable each recurrent unit to adaptively capture dependencies of different time scales. 

Besides these proposed advanced transition functions such as LSTMs and GRUs, multi-layer and bi-directional recurrent structure can increase the model capacity and flexibility. As shown in Figure \ref{illus_rnnvar}.(b), multi-layer structure can enable the hidden output of one recurrent layer to be propagated through time and used as the input data to the next recurrent layer. And the bidirectional recurrent structure is able to process the sequence data in two directions including forward and backward ways with two separate hidden layers, which is illustrated in Figure \ref{illus_rnnvar}.(c). The following equations define the corresponding hidden layer function and the $\rightarrow$ and $\leftarrow$ denote forward and backward process, respectively.    
\begin{align}
	\begin{split}\label{birnn}
\overrightarrow{\mathbf{h}}_t &= \overrightarrow{\mathbb{H}}(\mathbf{x} _{t}, \overrightarrow{\mathbf{h}}_{t-1}),\\
\overleftarrow{\mathbf{h}}_t &= \overleftarrow{\mathbb{H}}(\mathbf{x} _{t}, \overleftarrow{\mathbf{h}}_{t+1}).\\
\end{split}
\end{align} 
Then, the final vector $\mathbf{h}^{T}$ is the concatenated vector of the outputs of forward and backward processes as follows:
\begin{equation}\label{finalvec}
 \mathbf{h}_{T}= \overrightarrow{\mathbf{h}}_{T}\oplus \overleftarrow{\mathbf{h}}_{1}
\end{equation}

\section{Applications of Deep learning in machine health monitoring}\label{S3}
The conventional multilayer perceptron (MLP) has been applied in the field of machine health monitoring for many years \cite{4084702,li2000neural,samanta2003artificial,aminian2000neural}. The deep learning techniques have recently been applied to a large number of machine health monitoring systems. The layer-by-layer pretraining of deep neural network (DNN) based on Auto-encoder or RBM can facilitate the training of DNN and improve its discriminative power to characterize machinery data. Convolution neural network and recurrent neural networks provide more advanced and complex compositions mechanism to learn representation from machinery data. In these DL-based MHMS systems, the top layer normally represents the targets. For diagnosis where targets are discrete values, softmax layer is applied. For prognosis with continuous targets, liner regression layer is added. What is more, the end-to-end structure enables DL-based MHMS to be constructed with less human labor and expert knowledge, therefore these models are not limited to specific machine specific or domain. In the following, a brief survey of DL-based MHMS are presented in these above four DL architectures: AE, RBM, CNN and RNN.
\subsection{AE and its variants for machine health monitoring}
AE models, especially stacked DA, can learn high-level representations from machinery data in an automatic way. Sun et al. proposed a one layer AE-based neural network to classify induction motor faults \cite{sun2016sparse}. Due to the limited size of training data, they focused on how to prevent overfiting. Not only the number of hidden layer was set to 1, but also dropout technique that masks portions of output neurons randomly was applied on the hidden layer. However, most of proposed models are based on deep architectures by stacking multiple auto-encoders. Lu et al. presented a detailed empirical study of stacked denoising autoencoders with three hidden layers for fault diagnosis of rotary machinery components \cite{lu2017fault}. Specifically, in their experiments including single working condition and cross working conditions, the effectiveness of the receptive input size, deep architecture, sparsity constraint and denosing operation in the SDA model were evaluated. In \cite{tao2015bearing}, different structures of a two-layer SAE-based DNN were designed by varying hidden layer size and its masking probability, and evaluated for their performances in fault diagnosis.  

\begin{figure}
  \centering
  \includegraphics[width=0.4\textwidth]{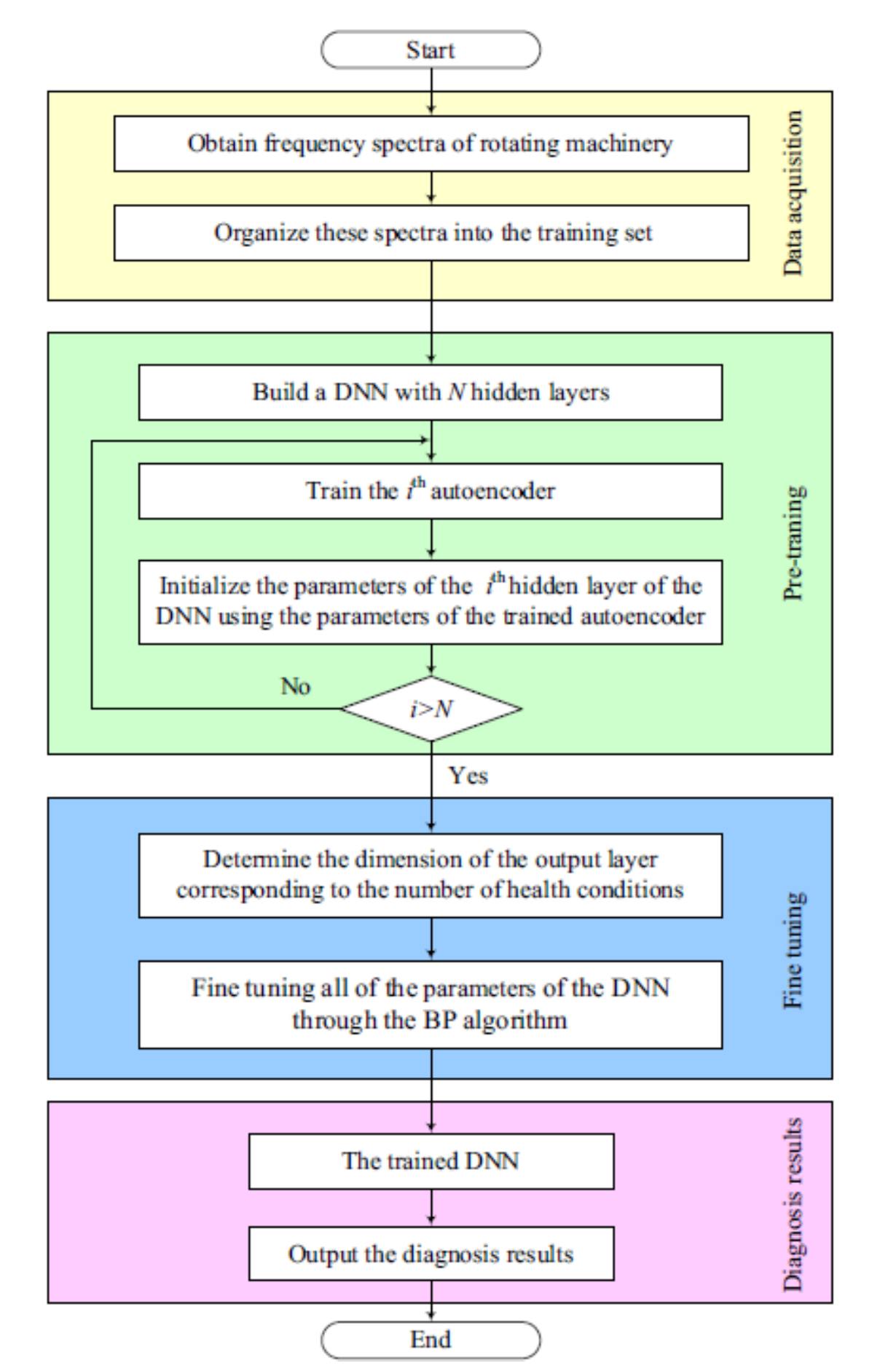}
  \caption{Illustrations of the proposed SAE-DNN for rotating machinery diagnosis in \cite{jia2016deep}.}
\label{illus_saeapp}
\end{figure}

In these above works, the input features to AE models are raw sensory time-series. Therefore, the input dimensionality is always over hundred, even one thousand. The possible high dimensionality may lead to some potential concerns such as heavy computation cost and overfiting caused by huge model parameters. Therefore, some researchers focused on AE models built upon features extracted from raw input. Jia et al. fed the frequency spectra of time-series data into SAE for rotating machinery diagnosis \cite{jia2016deep}, considering the frequency spectra is able to demonstrate how their constitutive components are distributed with discrete frequencies and may be more discriminative over the health conditions of rotating machinery. The corresponding framework proposed by Jia et al. is shown in Figure \ref{illus_saeapp}. Tan et al. used digital wavelet frame and nonlinear soft threshold method to process the vibration signal and built a SAE on the preprocessed signal for roller bearing fault diagnosis \cite{junbo2015fault}. Zhu et al. proposed a SAE-based DNN for hydraulic pump fault diagnosis with input as frequency domain features after Fourier transform \cite{7494195}. In experiments, \textit{ReLU} activation and dropout technique were analyzed and experimental results have shown to be effective in preventing gradient vanishing and overfiting. In the work presented in \cite{liu2016rolling}, the normalized spectrogram generated by STFT of sound signal was fed into two-layers SAE-based DNN for rolling bearing fault diagnosis. Galloway et al. built a two layer SAE-based DNN on spectrograms generated from raw vibration data for tidal turbine vibration fault diagnosis \cite{galloway2016diagnosis}. A SAE-based DNN with input as principal components of data extracted by principal component analysis was proposed for spacecraft fault diagnosis in \cite{li2015study}. Multi-domain statistical features including time domain features, frequency domain features and time-frequency domain features were fed into the SAE framework, which can be regarded as one kind of feature fusion \cite{guo2016multifeatures}. Similarly, Verma et al. also used these three domains features to fed into a SAE-based DNN for fault diagnosis of air compressors \cite{verma2013intelligent}. 

Except that these applied multi-domain features, multi-sensory data are also addressed by SAE models. Reddy utilized SAE to learn representation on raw time series data from multiple sensors for anomaly detection and fault disambiguation in flight data. To address multi-sensory data, synchronized windows were firstly traversed over multi-modal time series with overlap, and then windows from each sensor were concatenated as the input to the following SAE \cite{kk2016}. In \cite{li2017ae}, SAE was leveraged for multi-sensory data fusion and the followed DBN was adopted for bearing fault diagnosis, which achieved promising results. The statistical features in time domain and frequency domain extracted from the vibration signals of different sensors were adopted as input to a two-layer SAE with sparsity constraint neural networks. And the learned representation were fed into a deep belief network for pattern classification. 

In addition, some variants of the conventional SAE were proposed or introduced for machine health monitoring. In \cite{thirukovalluru2016generating}, Thirukovalluru et al. proposed a two-phase framework that SAE only learn representation and other standard classifiers such as SVM and random forest perform classification. Specifically, in SAE module, handcrafted features based on FFT and WPT were fed into SAE-based DNN. After pre-training and supervised fine-tuning which includes two separated procedures: softmax-based and Median-based fine-tuning methods, the extensive experiments on five datasets including air compressor monitoring, drill bit monitoring, bearing fault monitoring and steel plate monitoring have demonstrated the generalization capability of DL-based machine health monitoring systems. Wang et al. proposed a novel continuous sparse auto-encoder (CSAE) as an unsupervised feature learning for transformer fault recognition \cite{wang2016transformer}. Different from conventional sparse AE, their proposed CSAE added the stochastic unit into activation function of each visible unit as:
\begin{equation}\label{case}
s_{j}= \varphi_j(\sum_i{w_{ij}x{i}}+a_i+\sigma N_j(0,1))
\end{equation}
where $s_j$ is the output corresponding to the input $x_i$, $w_{ij}$ and $a_i$ denote model parameters, $\varphi_j$ represents the activation function and the last term $\sigma N_j(0,1))$ is the added stochastic unit, which is a zero-mean Gaussian with variance $\sigma^2$. The incorporation of stochastic unit is able to change the gradient direction and prevent over-fitting. Mao et al. adopted a variant of AE named Extreme Learning Machine-based auto-encoder for bearing fault diagnosis, which is more efficient than conventional SAE models without sacrificing accuracies in fault diagnosis \cite{mao2016bearing}. Different from AE that is trained via back-propagation, the transformation in encoder phase is randomly generated and the one in decoder phase is learned in a single step via least-squares fit \cite{6733226}.

In addition, Lu et al. focused on the visualization of learned representation by a two-layer SAE-based DNN, which provides a novel view to evaluate the DL-based MHMS \cite{lu2015novel}. In their paper, the discriminative power of learned representation can be improved with the increasing of layers.

\subsection{RBM and its variants for machine health monitoring}
In the section, some work focused on developing RBM and DBM to learn representation from machinery data. Most of works introduced here are based on deep belief networks (DBN) that can pretrain a deep neural network (DNN).

In \cite{deutschusing}, a RBM based method for bearing remaining useful life (RUL) prediction was proposed. Linear regression layer was added at the top of RBM after pretraining to predict the future root mean square (RMS) based on a lagged time series of RMS values. Then, RUL was calculated by using the predicted RMS and the total time of the bearing's life. Liao et al. proposed a new RBM for representation learning to predict RUL of machines \cite{liao2016enhanced}. In their work, a new regularization term modeling the trendability of the hidden nodes was added into the training objective function of RBM. Then, unsupervised self-organizing map algorithm (SOM) was applied to transforming the representation learned by the enhanced RBM to one scale named health value. Finally, the health value was used to predict RUL via a similarity-based life prediction algorithm. In \cite{li2015multimodal}, a multi-modal deep support vector classification approach was proposed for fault diagnosis of gearboxes. Firstly, three modalities features including time, frequency and time-frequency ones were extracted from vibration signals. Then, three Gaussian-Bernoulli deep Boltzmann machines (GDBMS) were applied to addressing the above three modalities, respectively. In each GDBMS, the softmax layer was used at the top. After the pretraining and the fine-tuning processes, the probabilistic outputs of the softmax layers from these three GDBMS were fused by a support vector classification (SVC) framework to make the final prediction. Li et al. applied one GDBMS directly on the concatenation feature consisting of three modalities features including time, frequency and time-frequency ones and stacked one softmax layer on top of GDBMS to recognize fault categories \cite{li2016fault}. Li et al. adopted a two-layers DBM to learn deep representations of the statistical parameters of the wavelet packet transform (WPT) of raw sensory signal for gearbox fault diagnosis \cite{li2016gearbox}. In this work focusing on data fusion, two DBMs were applied on acoustic and vibratory signals and random forest was applied to fusing the representations learned by these two DBMs. 

\begin{figure}
  \centering
  \includegraphics[width=0.5\textwidth]{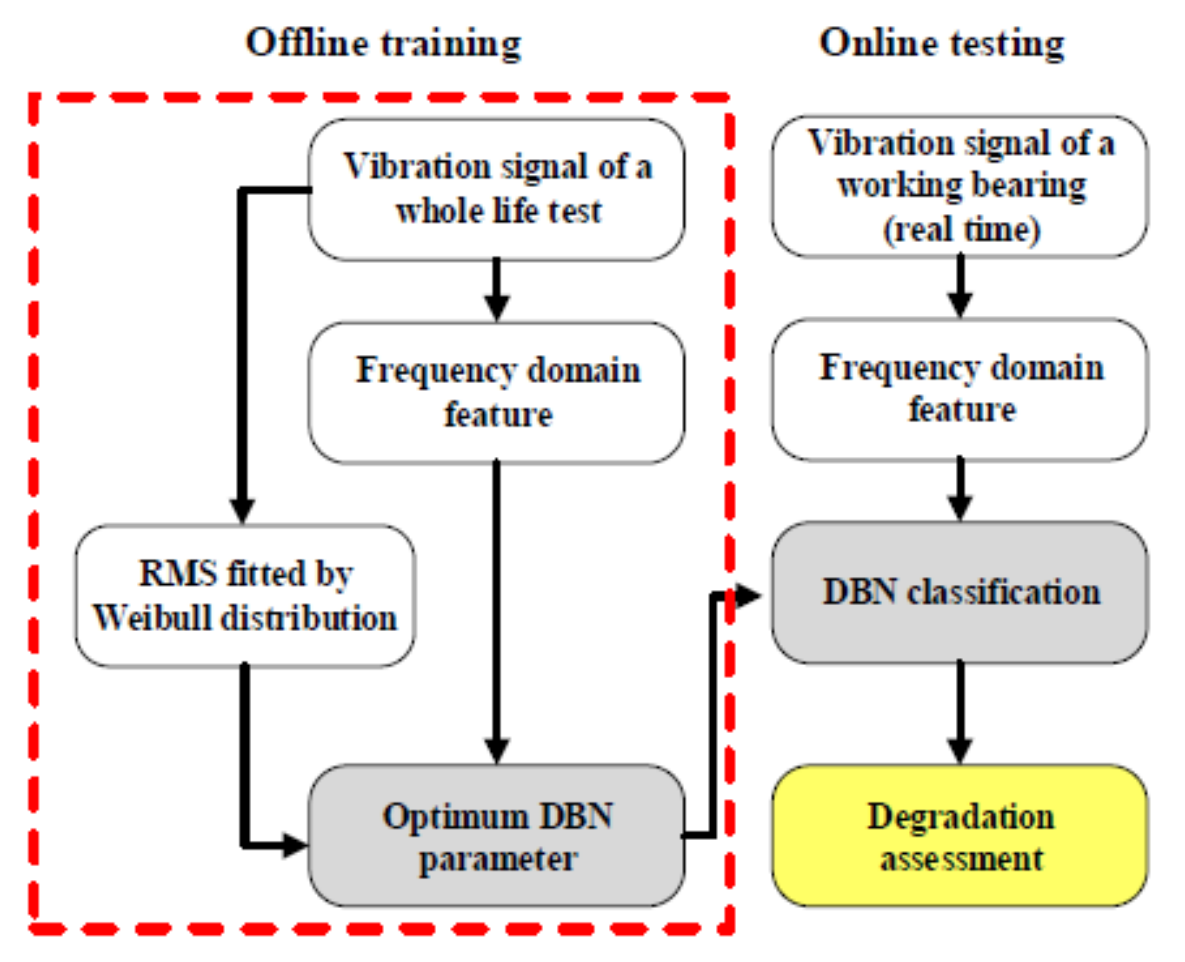}
  \caption{Illustrations of the proposed DBN-DNN for assessment of bearing degration in \cite{mengma2016}.}
\label{illus_dbnapp}
\end{figure}

Making use of DBN-based DNN, Ma et al. presented this framework for degradation assessment under a bearing accelerated life test \cite{mengma2016}. The statistical feature, root mean square (RMS) fitted by Weibull distribution that can avoid areas of fluctuation of the statistical parameter and the frequency domain features were extracted as raw input. To give a clear illustration, the framework in \cite{mengma2016} is shown in Figure \ref{illus_dbnapp}. Shao et al. proposed DBN for induction motor fault diagnosis with the direct usage of vibration signals as input \cite{shaoss2016}. Beside the evaluation of the final classification accuracies, t-SNE algorithm was adopted to visualize the learned representation of DBN and outputs of each layer in DBN. They found the addition of hidden layer can increase the discriminative power in the learned representation. Fu et al. employed deep belief networks for cutting states monitoring \cite{fu2015analysis}. In the presented work, three different feature sets including raw vibration signal, Mel-frequency cepstrum coefficient (MFCC) and wavelet features were fed into DBN as three corresponding different inputs, which were able to achieve robust comparative performance on the raw vibration signal without too much feature engineering. Tamilselvan et al. proposed a multi-sensory DBN-based health state classification model. The model was verified in benchmark classification problems and two health diagnosis applications including aircraft engine health diagnosis and electric power transformer health diagnosis \cite{tamilselvan2013failure,tamilselvan2012deep}. Tao et al. proposed DBN based multisensor information fusion scheme for bearing fault diagnosis \cite{tao2016bearing}. Firstly, 14 time-domain statistical features extracted from three vibration signals acquired by three sensors were concatenated together as an input vector to DBM model. During pre-training, a predefined threshold value was introduced to determine its iteration number. In \cite{chen2015multi}, a feature vector consisting of load and speed measure, time domain features and frequency domain features was fed into DBN-based DNN for gearbox fault diagnosis. In the work of \cite{gan2016construction}, Gan et al. built a hierarchical diagnosis network for fault pattern recognition of rolling element bearings consisting of two consecutive phases where the four different fault locations (including one health state) were firstly identified and then discrete fault severities in each fault condition were classified. In each phases, the frequency-band energy features generated by WPT were fed into DBN-based DNN for pattern classification. In \cite{oh2016dbn}, raw vibration signals were pre-processed to generate 2D image based on omnidirectional regeneration (ODR) techniques and then, histogram of original gradients (HOG) descriptor was applied on the generated image and the learned vector was fed into DBN for automatic diagnosis of journal bearing rotor systems. Chen et al. proposed an ensemble of DBNs with multi-objective evolutionary optimization on decomposition algorithm (MOEA/D) for fault diagnosis with multivariate sensory data \cite{zhang2015deep}. DBNs with different architectures can be regarded as base classifiers and MOEA/D was introduced to adjust the ensemble weights to achieve a trade-off between accuracy and diversity. Chen et al. then extended this above framework for one specific prognostics task: the RUL estimation of the mechanical system \cite{zhang2016multiobjective}.

\subsection{CNN for machine health monitoring}
In some scenarios, machinery data can be presented in a 2D format such as time-frequency spectrum, while in some scenarios, they are in a 1D format, i.e., time-series. Therefore, CNNs models are able to learn complex and robust representation via its convolutional layer. Intuitively, filters in convolutional layers can extract local patterns in raw data and stacking these convolutional layers can further build complex patterns. Janssens et al. utilized a 2D-CNN model for four categories rotating machinery conditions recognition, whose input is DFT of two accelerometer signals from two sensors that are placed perpendicular to each other. Therefore, the height of input is the number of sensors. The adopted CNN model contains one convolutional layer and a fully connected layer. Then, the top softmax layer is adopted for classification \cite{janssens2016convolutional}. In \cite{babu2016deep}, Babu et al. built a 2D deep convolution neural network to predict the RUL of system based on normalized-variate time series from sensor signals, in which one dimension of the 2D input is number of sensors as the setting reported in \cite{janssens2016convolutional}. In their model, average pooling is adopted instead of max pooling. Since RUL is a continuous value, the top layer was linear regression layer. Ding et al. proposed a deep Convolutional Network (ConvNet) where wavelet packet energy (WPE) image were used as input for spindle bearing fault diagnosis \cite{he2017cnn}. To fully discover the hierarchical representation, a multiscale layer was added after the last convolutional layer, which concatenates the outputs of the last convolutional layer and the ones of the previous pooling layer. Guo et al. proposed a hierarchical adaptive deep convolution neural network (ADCNN) \cite{guo2016hierarchical}. Firstly, the input time series data as a signal-vector was transformed into a $32\times32$ matrix, which follows the typical
input format adopted by LeNet \cite{lecun1998gradient}. In addition, they designed a hierarchical framework to recognize fault patterns and fault size. In the fault pattern decision module, the first ADCNN was adopted to recognize fault type. In the fault size evaluation layer, based on each fault type, ADCNN with the same structure was used to predict fault size. Here, the classification mechanism is still used. The predicted value $f$ is defined as the probability summation of the typical fault sizes as follows:
\begin{equation}\label{probssum}
f= \sum_{j=1}^{c}{a_{j}p_{j}}
\end{equation}
where $[p_1,\dots,p_c]$ is produced by the top softmax layer, which denote the probability score that each sample belongs to each class size and $a_j$ is the fault size corresponding to the $j$-th fault size. In \cite{wang2016}, an enhanced CNN was proposed for machinery fault diagnosis. To pre-process vibration data, morlet wavelet was used to decompose the vibration signal and obtain wavelet scaleogram. Then, bilinear interpolation was used to rescale the scaleogram into a grayscale image with a size of $32\times32$. In addition, the adaptation of ReLU and dropout both boost the model's diagnosis performance. Chen et al. adopted a 2D-CNN for gearbox fault diagnosis, in which the input matrix with a size of $16\times16$ for CNN is reshaped by a vector containing 256 statistic features including RMS values, standard
deviation, skewness, kurtosis, rotation frequency, and applied load \cite{chen2015gearbox}. In addition, 11 different structures of CNN were evaluated empirically in their experiments. Weimer et al. did a comprehensive study of various design configurations of deep CNN for visual defect detection \cite{weimer2016design}. In one specific application: industrial optical inspection, two directions of model configurations including depth (addition of conv-layer) and width (increase of number filters) were investigated. The optimal configuration verified empirically has been presented in Table \ref{compare_cnnmhs}. In \cite{dongsmall}, CNN was applied in the field of diagnosing the early small faults of front-end controlled wind generator (FSCWG) that the $784\times784$ input matrix consists of vibration data of generator input shaft (horizontal) and vibration data of generator output shaft (vertical) in time scale.

\begin{figure}
  \centering
  \includegraphics[width=0.5\textwidth]{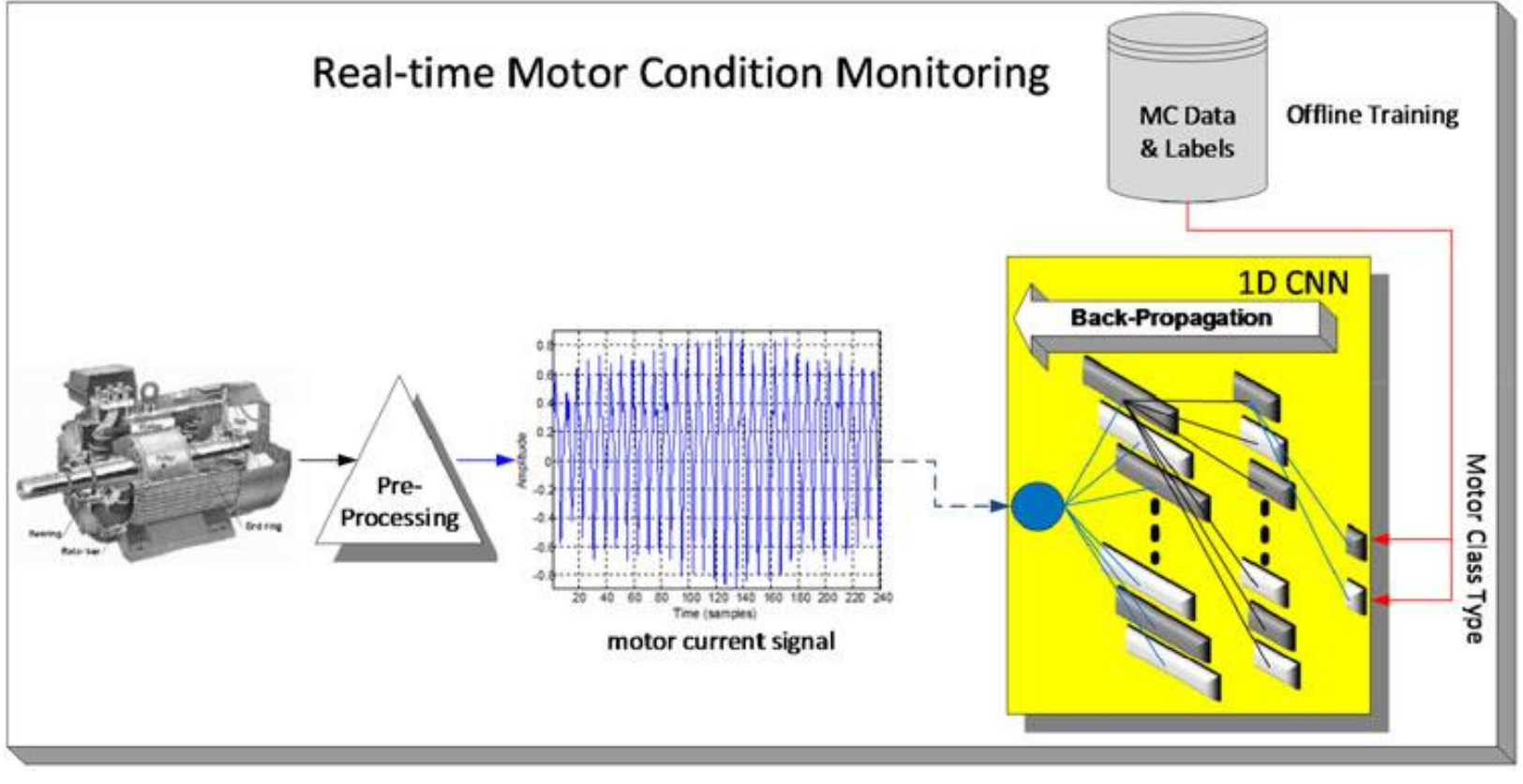}
  \caption{Illustrations of the proposed 1D-CNN for real-time motor Fault Detection in \cite{ince2016real}.}
\label{illus_cnnapp}
\end{figure}

As reviewed in our above section \ref{intro_cnn}, CNN can also be applied to 1D time series signal and the corresponding operations have been elaborated. In \cite{ince2016real}, the 1D CNN was successfully developed on raw time series data for motor fault detection, in which feature extraction and classification were integrated together. The corresponding framework has been shown in Figure \ref{illus_cnnapp}. Abdeljaber et al. proposed 1D CNN on normalized vibration signal, which can perform vibration-based damage detection and localization of the structural damage in real-time. The advantage of this approach is its ability to extract optimal damage-sensitive features automatically from the raw acceleration signals, which does not need any additional preporcessing or signal processing approaches \cite{abdeljaber2017real}.  

To present an overview about all these above CNN models that have been successfully applied in the area of MHMS, their architectures have been summarized in Table \ref{compare_cnnmhs}. To explain the used abbreviation, the structure of CNN applied in Weimer's work \cite{weimer2016design} is denoted as $\textnormal{Input}[32\times32]-64\textnormal{C}[3\times3]2-64\textnormal{P}[2\times2]-128\textnormal{C}[3\times3]3-128\textnormal{P}[2\times2]-\textnormal{FC}[1024-1024]2$. It means the input 2D data is $32\times32$ and the CNN firstly applied 2 convolutional layers with the same design that the filter number is 64 and the filter size is $3\times3$, then stacked one max-pooling layer whose pooling size is $2\times2$, then applied 3 convolutional layers with the same design that the filter number is 128 and the filer size is $3\times3$, then applied a pooling layer whose pooling size is $2\times2$, and finally adopted two fully-connected layers whose hidden neuron numbers are both 1024. It should be noted that the size of output layer is not given here, considering it is task-specific and usually set to be the number of categories.

\begin{table*}
\centering
\caption{Summary on configurations of CNN-based MHMS. The symbol \textit{Input}, \textit{C}, \textit{P} and \textit{FC} denote the raw input, convolutional layer, pooling layer and Fully-connected layer, respectively.}
\begin{tabular}{c|c|c} \hline
&Proposed Models&Configurations of CNN Structures\\ \hline
\multirow{8}{*}{2D CNN} 
 &\textit{Janssens's work \cite{janssens2016convolutional}}&$\textnormal{Input}[5120\times2]-32\textnormal{C}[64\times2]-\textnormal{FC}[200]$\\
           &  \textit{Babu's work \cite{babu2016deep}}& $\textnormal{Input}[27\times15]-8\textnormal{C}[27\times4]-8\textnormal{P}[1\times2]-14\textnormal{C}[1\times3]-14\textnormal{P}[1\times2]$\\
                     &  \textit{Ding's work \cite{he2017cnn}}& $\textnormal{Input}[32\times32]-20\textnormal{C}[7\times7]-20\textnormal{P}[2\times2]-10\textnormal{C}[6\times6]-10\textnormal{P}[2\times2]-6\textnormal{P}[2\times2]-\textnormal{FC}[185-24]$\\
 &\textit{Guo's work \cite{guo2016hierarchical}}&$\textnormal{Input}[32\times32]-5\textnormal{C}[5\times5]-5\textnormal{P}[2\times2]-10\textnormal{C}[5\times5]-
10\textnormal{P}[2\times2]-10\textnormal{C}[2\times2]-10\textnormal{P}[2\times2]-\textnormal{FC}[100]-\textnormal{FC}[50]$ \\
 &\textit{Wang's work \cite{wang2016}}& $\textnormal{Input}[32\times32]-64\textnormal{C}[3\times3]-64\textnormal{P}[2\times2]-64\textnormal{C}[4\times4]-
64\textnormal{P}[2\times2]-128\textnormal{C}[3\times3]-128\textnormal{P}[2\times2]-\textnormal{FC}[512]$ \\
 &\textit{Chen's work \cite{chen2015gearbox}}& $\textnormal{Input}[16\times16]-8\textnormal{C}[5\times5]-8\textnormal{P}[2\times2]$ \\
  &\textit{Weimer's work \cite{weimer2016design}}& $\textnormal{Input}[32\times32]-64\textnormal{C}[3\times3]2-64\textnormal{P}[2\times2]-128\textnormal{C}[3\times3]3-128\textnormal{P}[2\times2]-\textnormal{FC}[1024-1024]$ \\
   &\textit{Dong's work \cite{dongsmall}}& $\textnormal{Input}[784\times784]-12\textnormal{C}[10\times10]-12\textnormal{P}[2\times2]-24\textnormal{C}[10\times10]-24\textnormal{P}[2\times2]-\textnormal{FC}[200]$ \\ \hlinew{1pt}
   \multirow{2}{*}{1D CNN}
       &  \textit{Ince's work \cite{ince2016real}}& $\textnormal{Input}[240]-60\textnormal{C}[9]-60\textnormal{P}[4]-40\textnormal{C}[9]-40\textnormal{P}[4]-40\textnormal{C}[9]-40\textnormal{P}[4]-\textnormal{FC}[20]$ \\
               &  \textit{Abdeljaber's work \cite{abdeljaber2017real}}& $\textnormal{Input}[128]-64\textnormal{C}[41]-64\textnormal{P}[2]-32\textnormal{C}[41]-32\textnormal{P}[2]-\textnormal{FC}[10-10]$ 
               \\
   \hline
\end{tabular}
\label{compare_cnnmhs}
\end{table*}

\subsection{RNN for machine health monitoring}
\begin{figure*}
  \centering
  \includegraphics[width=1.0\textwidth]{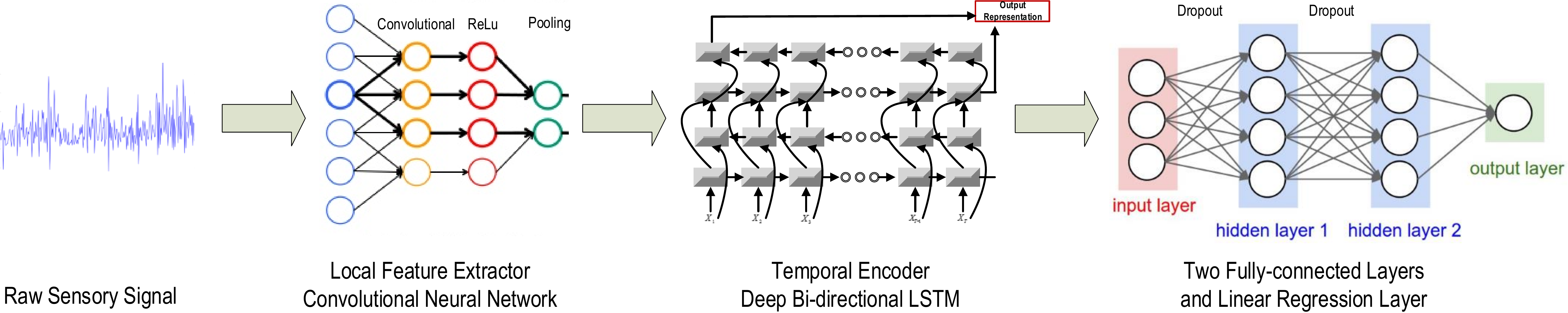}
  \caption{Illustrations of the proposed Convolutional Bi-directional Long Short-Term Memory Networks in \cite{ruiconvlstm}.}
\label{illus_convlstm}
\end{figure*}

The majority of machinery data belong to sensor data, which are in nature time series. RNN models including LSTM and GRU have emerged as one kind of popular architectures to handle sequential data with its ability to encode temporal information. These advanced RNN models have been proposed  to relief the difficulty of training in vanilla RNN and applied in machine health monitoring recently. In \cite{7748035}, Yuan et al. investigated three RNN models including vanilla RNN, LSTM and GRU models for fault diagnosis and prognostics of aero engine. They found these advanced RNN models LSTM and GRU models outperformed vanilla RNN. Another interesting observation was the ensemble model of the above three RNN variants did not boost the performance of LSTM. Zhao et al. presented an empirical evaluation of LSTMs-based machine health monitoring system in the tool wear test \cite{lstmrui}. The applied LSTM model encoded the raw sensory data into embeddings and predicted the corresponding tool wear. Zhao et al. further designed a more complex deep learning model combining CNN and LSTM named Convolutional Bi-directional Long Short-Term Memory Networks (CBLSTM) \cite{ruiconvlstm}. As shown in Figure \ref{illus_convlstm}, CNN was used to extract robust local features from the sequential input, and then bi-directional LSTM was adopted to encode temporal information on the sequential output of CNN. Stacked fully-connected layers and linear regression layer were finally added to predict the target value. In tool wear test, the proposed model was able to outperform several state-of-the-art baseline methods including conventional LSTM models. In \cite{malhotra2016multi}, Malhotra proposed a very interesting structure for RUL prediction. They designed a LSTM-based encoder-decoder structure, which LSTM-based encoder firstly transforms a multivariate input sequence to a fixed-length vector and then, LSTM decoder uses the vector to produce the target sequence. When it comes to RUL prediction, their assumptions lies that the model can be firstly trained in raw signal corresponding to normal behavior in an unsupervised way. Then, the reconstruction error can be used to compute health index (HI), which is then used for RUL estimation. It is intuitive that the large reconstruction error corresponds to a more unhealthy machine condition.

\section{Summary and Future Directions}\label{S4}
In this paper, we have provided a systematic overview of the state-of-the-art DL-based MHMS. Deep learning, as a sub-field of machine learning, is serving as a bridge between big machinery data and data-driven MHMS. Therefore, within the past four years, they have been applied in various machine health monitoring tasks. These proposed DL-based MHMS are summarized according to four categories of DL architecture as: Auto-encoder models, Restricted Boltzmann Machines models, Convolutional Neural Networks and Recurrent Neural Networks. Since the momentum of the research of DL-based MHMS is growing fast, we hope the messages about the capabilities of these DL techniques, especially representation learning for complex machinery data and target prediction for various machine health monitoring tasks, can be conveyed to readers. Through these previous works, it can be found that DL-based MHMS do not require extensive human labor and expert knowledge, i.e., the end-to-end structure is able to map raw machinery data to targets. Therefore, the application of deep learning models are not restricted to specific kinds of machines, which can be a general solution to address the machine health monitoring problems. Besides, some research trends and potential future research directions are given as follows:
\renewcommand{\labelenumii}{\Roman{enumii}}
\begin{enumerate}
\item[*] \textbf{Open-source Large Dataset}: Due to the huge model complexity behind DL methods, the performance of DL-based MHMS heavily depends on the scale and quality of datasets. On other hand, the depth of DL model is limited by the scale of datasets. As a result, the benchmark CNN model for image recognition has 152 layers, which can be supported by the large dataset ImageNet containing over ten million annotated images \cite{he2015deep,imagenet_cvpr09}. In contrast, the proposed DL models for MHMS may stack up to 5 hidden layers. And the model trained in such kind of large datasets can be the model initialization for the following specific task/dataset. Therefore, it is meaningful to design and publish large-scale machinery datasets.

\item[*] \textbf{Utilization of Domain Knowledge}: deep learning is not a skeleton key to all machine health monitoring problems. Domain knowledge can contribute to the success of applying DL models on machine health monitoring. For example, extracting discriminative features can reduce the size of the followed DL models and appropriate task-specific regularization term can boost the final performance \cite{liao2016enhanced}.     

\item[*] \textbf{Model and Data Visualization}: deep learning techniques, especially deep neural networks, have been regarded as black boxes models, i.e., their inner computation mechanisms are unexplainable. Visualization of the learned representation and the applied model can offer some insights into these DL models, and then these insights achieved by this kind of interaction can facilitate the building and configuration of DL models for complex machine health monitoring problems. Some visualization techniques have been proposed including t-SNE model for high dimensional data visualization \cite{maaten2008visualizing} and visualization of the activations produced by each layer and features at each layer of a DNN via regularized optimization \cite{yosinski2015understanding}.

\item[*] \textbf{Transferred Deep Learning}: Transfer learning tries to apply knowledge learned in one domain to a different but related domain \cite{pan2010survey}. This research direction is meaningful in machine health monitoring, since some machine health monitoring problems have sufficient training data while other areas lack training data. The machine learning models including DL models trained in one domain can be transferred to the other domain. Some previous works focusing on transferred feature extraction/dimensionality reduction have been done \cite{shen2015bearing,7542845}. In \cite{lu2016deep}, a Maximum Mean Discrepancy (MMD) measure evaluating the discrepancy between source and target domains was added into the target function of deep neural networks. 

\item[*] \textbf{Imbalanced Class}: The class distribution of machinery data in real life normally follows a highly-skewed one, in which most data samples belong to few categories. For example, the number of fault data is much less than the one of health data in fault diagnosis. Some enhanced machine learning models including SVM and ELM have been proposed to address this imbalanced issue in machine health monitoring \cite{mao2017online,duan2016new}. Recently, some interesting methods investigating the application of deep learning in imbalanced class problems have been developed, including CNN models with class resampling or cost-sensitive training \cite{huang2016learning} and the integration of boot strapping methods and CNN model \cite{yan2015deep}. 
\end{enumerate}

It is believed that deep learning will have a more and more prospective future impacting machine health monitoring, especially in the age of big machinery data.

\section*{Acknowledgment}
This work has been supported in part by the National Natural Science Foundation of China (51575102).

\ifCLASSOPTIONcaptionsoff
  \newpage
\fi

\bibliographystyle{IEEEtran}
\bibliography{reference,ae_ref,cnn_ref,rbm_ref,rnn_ref,tl_ref}

\begin{thebibliography}{100}
\providecommand{\url}[1]{#1}
\csname url@samestyle\endcsname
\providecommand{\newblock}{\relax}
\providecommand{\bibinfo}[2]{#2}
\providecommand{\BIBentrySTDinterwordspacing}{\spaceskip=0pt\relax}
\providecommand{\BIBentryALTinterwordstretchfactor}{4}
\providecommand{\BIBentryALTinterwordspacing}{\spaceskip=\fontdimen2\font plus
\BIBentryALTinterwordstretchfactor\fontdimen3\font minus
  \fontdimen4\font\relax}
\providecommand{\BIBforeignlanguage}[2]{{%
\expandafter\ifx\csname l@#1\endcsname\relax
\typeout{** WARNING: IEEEtran.bst: No hyphenation pattern has been}%
\typeout{** loaded for the language `#1'. Using the pattern for}%
\typeout{** the default language instead.}%
\else
\language=\csname l@#1\endcsname
\fi
#2}}
\providecommand{\BIBdecl}{\relax}
\BIBdecl

\bibitem{6748057}
S.~Yin, X.~Li, H.~Gao, and O.~Kaynak, ``Data-based techniques focused on modern
  industry: An overview,'' \emph{IEEE Transactions on Industrial Electronics},
  vol.~62, no.~1, pp. 657--667, Jan 2015.

\bibitem{jeschkeindustrial}
S.~Jeschke, C.~Brecher, H.~Song, and D.~B. Rawat, ``Industrial internet of
  things.''

\bibitem{lund2014worldwide}
D.~Lund, C.~MacGillivray, V.~Turner, and M.~Morales, ``Worldwide and regional
  internet of things (iot) 2014--2020 forecast: A virtuous circle of proven
  value and demand,'' \emph{International Data Corporation (IDC), Tech. Rep},
  2014.

\bibitem{li2000stochastic}
Y.~Li, T.~Kurfess, and S.~Liang, ``Stochastic prognostics for rolling element
  bearings,'' \emph{Mechanical Systems and Signal Processing}, vol.~14, no.~5,
  pp. 747--762, 2000.

\bibitem{oppenheimer2002physically}
C.~H. Oppenheimer and K.~A. Loparo, ``Physically based diagnosis and prognosis
  of cracked rotor shafts,'' in \emph{AeroSense 2002}.\hskip 1em plus 0.5em
  minus 0.4em\relax International Society for Optics and Photonics, 2002, pp.
  122--132.

\bibitem{yu2014model}
M.~Yu, D.~Wang, and M.~Luo, ``Model-based prognosis for hybrid systems with
  mode-dependent degradation behaviors,'' \emph{Industrial Electronics, IEEE
  Transactions on}, vol.~61, no.~1, pp. 546--554, 2014.

\bibitem{jardine2006review}
A.~K. Jardine, D.~Lin, and D.~Banjevic, ``A review on machinery diagnostics and
  prognostics implementing condition-based maintenance,'' \emph{Mechanical
  systems and signal processing}, vol.~20, no.~7, pp. 1483--1510, 2006.

\bibitem{ren2015faster}
S.~Ren, K.~He, R.~Girshick, and J.~Sun, ``Faster r-cnn: Towards real-time
  object detection with region proposal networks,'' in \emph{Advances in neural
  information processing systems}, 2015, pp. 91--99.

\bibitem{collobert2008unified}
R.~Collobert and J.~Weston, ``A unified architecture for natural language
  processing: Deep neural networks with multitask learning,'' in
  \emph{Proceedings of the 25th international conference on Machine
  learning}.\hskip 1em plus 0.5em minus 0.4em\relax ACM, 2008, pp. 160--167.

\bibitem{hinton2012deep}
G.~Hinton, L.~Deng, D.~Yu, G.~E. Dahl, A.-r. Mohamed, N.~Jaitly, A.~Senior,
  V.~Vanhoucke, P.~Nguyen, T.~N. Sainath \emph{et~al.}, ``Deep neural networks
  for acoustic modeling in speech recognition: The shared views of four
  research groups,'' \emph{IEEE Signal Processing Magazine}, vol.~29, no.~6,
  pp. 82--97, 2012.

\bibitem{leung2014deep}
M.~K. Leung, H.~Y. Xiong, L.~J. Lee, and B.~J. Frey, ``Deep learning of the
  tissue-regulated splicing code,'' \emph{Bioinformatics}, vol.~30, no.~12, pp.
  i121--i129, 2014.

\bibitem{888}
J.~Schmidhuber, ``Deep learning in neural networks: An overview,'' \emph{Neural
  Networks}, vol.~61, pp. 85--117, 2015, published online 2014; based on TR
  arXiv:1404.7828 [cs.NE].

\bibitem{lecun2015deep}
Y.~LeCun, Y.~Bengio, and G.~Hinton, ``Deep learning,'' \emph{Nature}, vol. 521,
  no. 7553, pp. 436--444, 2015.

\bibitem{raina2009large}
R.~Raina, A.~Madhavan, and A.~Y. Ng, ``Large-scale deep unsupervised learning
  using graphics processors,'' in \emph{Proceedings of the 26th annual
  international conference on machine learning}.\hskip 1em plus 0.5em minus
  0.4em\relax ACM, 2009, pp. 873--880.

\bibitem{hinton2007learning}
G.~E. Hinton, ``Learning multiple layers of representation,'' \emph{Trends in
  cognitive sciences}, vol.~11, no.~10, pp. 428--434, 2007.

\bibitem{widodo2007support}
A.~Widodo and B.-S. Yang, ``Support vector machine in machine condition
  monitoring and fault diagnosis,'' \emph{Mechanical systems and signal
  processing}, vol.~21, no.~6, pp. 2560--2574, 2007.

\bibitem{yan2005degradation}
J.~Yan and J.~Lee, ``Degradation assessment and fault modes classification
  using logistic regression,'' \emph{Journal of manufacturing Science and
  Engineering}, vol. 127, no.~4, pp. 912--914, 2005.

\bibitem{muralidharan2012comparative}
V.~Muralidharan and V.~Sugumaran, ``A comparative study of na{\"\i}ve bayes
  classifier and bayes net classifier for fault diagnosis of monoblock
  centrifugal pump using wavelet analysis,'' \emph{Applied Soft Computing},
  vol.~12, no.~8, pp. 2023--2029, 2012.

\bibitem{bengio2013representation}
Y.~Bengio, A.~Courville, and P.~Vincent, ``Representation learning: A review
  and new perspectives,'' \emph{IEEE transactions on pattern analysis and
  machine intelligence}, vol.~35, no.~8, pp. 1798--1828, 2013.

\bibitem{malhi2004pca}
A.~Malhi and R.~X. Gao, ``Pca-based feature selection scheme for machine defect
  classification,'' \emph{IEEE Transactions on Instrumentation and
  Measurement}, vol.~53, no.~6, pp. 1517--1525, 2004.

\bibitem{wang2016multisensory}
J.~Wang, J.~Xie, R.~Zhao, L.~Zhang, and L.~Duan, ``Multisensory fusion based
  virtual tool wear sensing for ubiquitous manufacturing,'' \emph{Robotics and
  Computer-Integrated Manufacturing}, 2016.

\bibitem{7517325}
J.~Wang, J.~Xie, R.~Zhao, K.~Mao, and L.~Zhang, ``A new probabilistic kernel
  factor analysis for multisensory data fusion: Application to tool condition
  monitoring,'' \emph{IEEE Transactions on Instrumentation and Measurement},
  vol.~65, no.~11, pp. 2527--2537, Nov 2016.

\bibitem{vincent2008extracting}
P.~Vincent, H.~Larochelle, Y.~Bengio, and P.-A. Manzagol, ``Extracting and
  composing robust features with denoising autoencoders,'' in \emph{Proceedings
  of the 25th international conference on Machine learning}.\hskip 1em plus
  0.5em minus 0.4em\relax ACM, 2008, pp. 1096--1103.

\bibitem{hinton2006fast}
G.~E. Hinton, S.~Osindero, and Y.-W. Teh, ``A fast learning algorithm for deep
  belief nets,'' \emph{Neural computation}, vol.~18, no.~7, pp. 1527--1554,
  2006.

\bibitem{salakhutdinov2009deep}
R.~Salakhutdinov and G.~E. Hinton, ``Deep boltzmann machines.'' in
  \emph{AISTATS}, vol.~1, 2009, p.~3.

\bibitem{sermanet2012convolutional}
P.~Sermanet, S.~Chintala, and Y.~LeCun, ``Convolutional neural networks applied
  to house numbers digit classification,'' in \emph{Pattern Recognition (ICPR),
  2012 21st International Conference on}.\hskip 1em plus 0.5em minus
  0.4em\relax IEEE, 2012, pp. 3288--3291.

\bibitem{funahashi1993approximation}
K.-i. Funahashi and Y.~Nakamura, ``Approximation of dynamical systems by
  continuous time recurrent neural networks,'' \emph{Neural networks}, vol.~6,
  no.~6, pp. 801--806, 1993.

\bibitem{Deng2014book}
\BIBentryALTinterwordspacing
L.~Deng and D.~Yu, ``Deep learning: Methods and applications,'' \emph{Found.
  Trends Signal Process.}, vol.~7, no. 3\&\#8211;4, pp. 197--387, Jun. 2014.
  [Online]. Available: \url{http://dx.doi.org/10.1561/2000000039}
\BIBentrySTDinterwordspacing

\bibitem{goodfellow2016deep}
I.~Goodfellow, Y.~Bengio, and A.~Courville, ``Deep learning,'' \emph{2015},
  2016.

\bibitem{bengio2007greedy}
Y.~Bengio, P.~Lamblin, D.~Popovici, H.~Larochelle \emph{et~al.}, ``Greedy
  layer-wise training of deep networks,'' \emph{Advances in neural information
  processing systems}, vol.~19, p. 153, 2007.

\bibitem{ng2011sparse}
A.~Ng, ``Sparse autoencoder,'' \emph{CS294A Lecture notes}, vol.~72, pp. 1--19,
  2011.

\bibitem{le1990handwritten}
B.~B. Le~Cun, J.~S. Denker, D.~Henderson, R.~E. Howard, W.~Hubbard, and L.~D.
  Jackel, ``Handwritten digit recognition with a back-propagation network,'' in
  \emph{Advances in neural information processing systems}.\hskip 1em plus
  0.5em minus 0.4em\relax Citeseer, 1990.

\bibitem{jarrett2009best}
K.~Jarrett, K.~Kavukcuoglu, Y.~Lecun \emph{et~al.}, ``What is the best
  multi-stage architecture for object recognition?'' in \emph{2009 IEEE 12th
  International Conference on Computer Vision}.\hskip 1em plus 0.5em minus
  0.4em\relax IEEE, 2009, pp. 2146--2153.

\bibitem{krizhevsky2012imagenet}
A.~Krizhevsky, I.~Sutskever, and G.~E. Hinton, ``Imagenet classification with
  deep convolutional neural networks,'' in \emph{Advances in neural information
  processing systems}, 2012, pp. 1097--1105.

\bibitem{abdel2012applying}
O.~Abdel-Hamid, A.-r. Mohamed, H.~Jiang, and G.~Penn, ``Applying convolutional
  neural networks concepts to hybrid nn-hmm model for speech recognition,'' in
  \emph{2012 IEEE international conference on Acoustics, speech and signal
  processing (ICASSP)}.\hskip 1em plus 0.5em minus 0.4em\relax IEEE, 2012, pp.
  4277--4280.

\bibitem{kim2014convolutional}
Y.~Kim, ``Convolutional neural networks for sentence classification,''
  \emph{arXiv preprint arXiv:1408.5882}, 2014.

\bibitem{jaeger2002tutorial}
H.~Jaeger, \emph{Tutorial on training recurrent neural networks, covering BPPT,
  RTRL, EKF and the" echo state network" approach}.\hskip 1em plus 0.5em minus
  0.4em\relax GMD-Forschungszentrum Informationstechnik, 2002.

\bibitem{giles1992learning}
C.~L. Giles, C.~B. Miller, D.~Chen, H.-H. Chen, G.-Z. Sun, and Y.-C. Lee,
  ``Learning and extracting finite state automata with second-order recurrent
  neural networks,'' \emph{Neural Computation}, vol.~4, no.~3, pp. 393--405,
  1992.

\bibitem{hochreiter1997long}
S.~Hochreiter and J.~Schmidhuber, ``Long short-term memory,'' \emph{Neural
  computation}, vol.~9, no.~8, pp. 1735--1780, 1997.

\bibitem{gers2000learning}
F.~A. Gers, J.~Schmidhuber, and F.~Cummins, ``Learning to forget: Continual
  prediction with lstm,'' \emph{Neural computation}, vol.~12, no.~10, pp.
  2451--2471, 2000.

\bibitem{gers2002learning}
F.~A. Gers, N.~N. Schraudolph, and J.~Schmidhuber, ``Learning precise timing
  with lstm recurrent networks,'' \emph{Journal of machine learning research},
  vol.~3, no. Aug, pp. 115--143, 2002.

\bibitem{cho2014learning}
K.~Cho, B.~Van~Merri{\"e}nboer, C.~Gulcehre, D.~Bahdanau, F.~Bougares,
  H.~Schwenk, and Y.~Bengio, ``Learning phrase representations using rnn
  encoder-decoder for statistical machine translation,'' \emph{arXiv preprint
  arXiv:1406.1078}, 2014.

\bibitem{chung2014empirical}
J.~Chung, C.~Gulcehre, K.~Cho, and Y.~Bengio, ``Empirical evaluation of gated
  recurrent neural networks on sequence modeling,'' \emph{arXiv preprint
  arXiv:1412.3555}, 2014.

\bibitem{4084702}
H.~Su and K.~T. Chong, ``Induction machine condition monitoring using neural
  network modeling,'' \emph{IEEE Transactions on Industrial Electronics},
  vol.~54, no.~1, pp. 241--249, Feb 2007.

\bibitem{li2000neural}
B.~Li, M.-Y. Chow, Y.~Tipsuwan, and J.~C. Hung, ``Neural-network-based motor
  rolling bearing fault diagnosis,'' \emph{IEEE transactions on industrial
  electronics}, vol.~47, no.~5, pp. 1060--1069, 2000.

\bibitem{samanta2003artificial}
B.~Samanta and K.~Al-Balushi, ``Artificial neural network based fault
  diagnostics of rolling element bearings using time-domain features,''
  \emph{Mechanical systems and signal processing}, vol.~17, no.~2, pp.
  317--328, 2003.

\bibitem{aminian2000neural}
M.~Aminian and F.~Aminian, ``Neural-network based analog-circuit fault
  diagnosis using wavelet transform as preprocessor,'' \emph{IEEE Transactions
  on Circuits and Systems II: Analog and Digital Signal Processing}, vol.~47,
  no.~2, pp. 151--156, 2000.

\bibitem{sun2016sparse}
W.~Sun, S.~Shao, R.~Zhao, R.~Yan, X.~Zhang, and X.~Chen, ``A sparse
  auto-encoder-based deep neural network approach for induction motor faults
  classification,'' \emph{Measurement}, vol.~89, pp. 171--178, 2016.

\bibitem{lu2017fault}
C.~Lu, Z.-Y. Wang, W.-L. Qin, and J.~Ma, ``Fault diagnosis of rotary machinery
  components using a stacked denoising autoencoder-based health state
  identification,'' \emph{Signal Processing}, vol. 130, pp. 377--388, 2017.

\bibitem{tao2015bearing}
S.~Tao, T.~Zhang, J.~Yang, X.~Wang, and W.~Lu, ``Bearing fault diagnosis method
  based on stacked autoencoder and softmax regression,'' in \emph{Control
  Conference (CCC), 2015 34th Chinese}.\hskip 1em plus 0.5em minus 0.4em\relax
  IEEE, 2015, pp. 6331--6335.

\bibitem{jia2016deep}
F.~Jia, Y.~Lei, J.~Lin, X.~Zhou, and N.~Lu, ``Deep neural networks: A promising
  tool for fault characteristic mining and intelligent diagnosis of rotating
  machinery with massive data,'' \emph{Mechanical Systems and Signal
  Processing}, vol.~72, pp. 303--315, 2016.

\bibitem{junbo2015fault}
T.~Junbo, L.~Weining, A.~Juneng, and W.~Xueqian, ``Fault diagnosis method study
  in roller bearing based on wavelet transform and stacked auto-encoder,'' in
  \emph{The 27th Chinese Control and Decision Conference (2015 CCDC)}.\hskip
  1em plus 0.5em minus 0.4em\relax IEEE, 2015, pp. 4608--4613.

\bibitem{7494195}
Z.~Huijie, R.~Ting, W.~Xinqing, Z.~You, and F.~Husheng, ``Fault diagnosis of
  hydraulic pump based on stacked autoencoders,'' in \emph{2015 12th IEEE
  International Conference on Electronic Measurement Instruments (ICEMI)},
  vol.~01, July 2015, pp. 58--62.

\bibitem{liu2016rolling}
H.~Liu, L.~Li, and J.~Ma, ``Rolling bearing fault diagnosis based on stft-deep
  learning and sound signals,'' \emph{Shock and Vibration}, vol. 2016, 2016.

\bibitem{galloway2016diagnosis}
G.~S. Galloway, V.~M. Catterson, T.~Fay, A.~Robb, and C.~Love, ``Diagnosis of
  tidal turbine vibration data through deep neural networks,'' 2016.

\bibitem{li2015study}
K.~Li and Q.~Wang, ``Study on signal recognition and diagnosis for spacecraft
  based on deep learning method,'' in \emph{Prognostics and System Health
  Management Conference (PHM), 2015}.\hskip 1em plus 0.5em minus 0.4em\relax
  IEEE, 2015, pp. 1--5.

\bibitem{guo2016multifeatures}
L.~Guo, H.~Gao, H.~Huang, X.~He, and S.~Li, ``Multifeatures fusion and
  nonlinear dimension reduction for intelligent bearing condition monitoring,''
  \emph{Shock and Vibration}, vol. 2016, 2016.

\bibitem{verma2013intelligent}
N.~K. Verma, V.~K. Gupta, M.~Sharma, and R.~K. Sevakula, ``Intelligent
  condition based monitoring of rotating machines using sparse auto-encoders,''
  in \emph{Prognostics and Health Management (PHM), 2013 IEEE Conference
  on}.\hskip 1em plus 0.5em minus 0.4em\relax IEEE, 2013, pp. 1--7.

\bibitem{kk2016}
v.~v. kishore~k. reddy, soumalya~sarkar and michael giering, ``Anomaly
  detection and fault disambiguation in large flight data: A multi-modal deep
  auto-encoder approach,'' in \emph{Annual conference of the prognostics and
  health management society, Denver, Colorado}, 2016.

\bibitem{li2017ae}
Z.~Chen and W.~Li, ``Multi-sensor feature fusion for bearing fault diagnosis
  using sparse auto encoder and deep belief network,'' \emph{IEEE Transactions
  on IM}, 2017.

\bibitem{thirukovalluru2016generating}
R.~Thirukovalluru, S.~Dixit, R.~K. Sevakula, N.~K. Verma, and A.~Salour,
  ``Generating feature sets for fault diagnosis using denoising stacked
  auto-encoder,'' in \emph{Prognostics and Health Management (ICPHM), 2016 IEEE
  International Conference on}.\hskip 1em plus 0.5em minus 0.4em\relax IEEE,
  2016, pp. 1--7.

\bibitem{wang2016transformer}
L.~Wang, X.~Zhao, J.~Pei, and G.~Tang, ``Transformer fault diagnosis using
  continuous sparse autoencoder,'' \emph{SpringerPlus}, vol.~5, no.~1, p.~1,
  2016.

\bibitem{mao2016bearing}
W.~Mao, J.~He, Y.~Li, and Y.~Yan, ``Bearing fault diagnosis with auto-encoder
  extreme learning machine: A comparative study,'' \emph{Proceedings of the
  Institution of Mechanical Engineers, Part C: Journal of Mechanical
  Engineering Science}, pp. 1--19, 2016.

\bibitem{6733226}
E.~Cambria, G.~B. Huang, L.~L.~C. Kasun, H.~Zhou, C.~M. Vong, J.~Lin, J.~Yin,
  Z.~Cai, Q.~Liu, K.~Li, V.~C.~M. Leung, L.~Feng, Y.~S. Ong, M.~H. Lim,
  A.~Akusok, A.~Lendasse, F.~Corona, R.~Nian, Y.~Miche, P.~Gastaldo, R.~Zunino,
  S.~Decherchi, X.~Yang, K.~Mao, B.~S. Oh, J.~Jeon, K.~A. Toh, A.~B.~J. Teoh,
  J.~Kim, H.~Yu, Y.~Chen, and J.~Liu, ``Extreme learning machines [trends
  controversies],'' \emph{IEEE Intelligent Systems}, vol.~28, no.~6, pp.
  30--59, Nov 2013.

\bibitem{lu2015novel}
W.~Lu, X.~Wang, C.~Yang, and T.~Zhang, ``A novel feature extraction method
  using deep neural network for rolling bearing fault diagnosis,'' in \emph{The
  27th Chinese Control and Decision Conference (2015 CCDC)}.\hskip 1em plus
  0.5em minus 0.4em\relax IEEE, 2015, pp. 2427--2431.

\bibitem{deutschusing}
J.~Deutsch and D.~He, ``Using deep learning based approaches for bearing
  remaining useful life prediction,'' 2016.

\bibitem{liao2016enhanced}
L.~Liao, W.~Jin, and R.~Pavel, ``Enhanced restricted boltzmann machine with
  prognosability regularization for prognostics and health assessment,''
  \emph{IEEE Transactions on Industrial Electronics}, vol.~63, no.~11, 2016.

\bibitem{li2015multimodal}
C.~Li, R.-V. Sanchez, G.~Zurita, M.~Cerrada, D.~Cabrera, and R.~E. V{\'a}squez,
  ``Multimodal deep support vector classification with homologous features and
  its application to gearbox fault diagnosis,'' \emph{Neurocomputing}, vol.
  168, pp. 119--127, 2015.

\bibitem{li2016fault}
C.~Li, R.-V. S{\'a}nchez, G.~Zurita, M.~Cerrada, and D.~Cabrera, ``Fault
  diagnosis for rotating machinery using vibration measurement deep statistical
  feature learning,'' \emph{Sensors}, vol.~16, no.~6, p. 895, 2016.

\bibitem{li2016gearbox}
C.~Li, R.-V. Sanchez, G.~Zurita, M.~Cerrada, D.~Cabrera, and R.~E. V{\'a}squez,
  ``Gearbox fault diagnosis based on deep random forest fusion of acoustic and
  vibratory signals,'' \emph{Mechanical Systems and Signal Processing},
  vol.~76, pp. 283--293, 2016.

\bibitem{mengma2016}
M.~Ma, X.~Chen, S.~Wang, Y.~Liu, and W.~Li, ``Bearing degradation assessment
  based on weibull distribution and deep belief network,'' in \emph{Proceedings
  of 2016 International Symposium of Flexible Automation (ISFA)}, 2016, pp.
  1--4.

\bibitem{shaoss2016}
S.~Shao, W.~Sun, P.~Wang, R.~X. Gao, and R.~Yan, ``Learning features from
  vibration signals for induction motor fault diagnosis,'' in \emph{Proceedings
  of 2016 International Symposium of Flexible Automation (ISFA)}, 2016, pp.
  1--6.

\bibitem{fu2015analysis}
Y.~Fu, Y.~Zhang, H.~Qiao, D.~Li, H.~Zhou, and J.~Leopold, ``Analysis of feature
  extracting ability for cutting state monitoring using deep belief networks,''
  \emph{Procedia CIRP}, vol.~31, pp. 29--34, 2015.

\bibitem{tamilselvan2013failure}
P.~Tamilselvan and P.~Wang, ``Failure diagnosis using deep belief learning
  based health state classification,'' \emph{Reliability Engineering \& System
  Safety}, vol. 115, pp. 124--135, 2013.

\bibitem{tamilselvan2012deep}
P.~Tamilselvan, Y.~Wang, and P.~Wang, ``Deep belief network based state
  classification for structural health diagnosis,'' in \emph{Aerospace
  Conference, 2012 IEEE}.\hskip 1em plus 0.5em minus 0.4em\relax IEEE, 2012,
  pp. 1--11.

\bibitem{tao2016bearing}
J.~Tao, Y.~Liu, and D.~Yang, ``Bearing fault diagnosis based on deep belief
  network and multisensor information fusion,'' \emph{Shock and Vibration},
  vol. 2016, 2016.

\bibitem{chen2015multi}
Z.~Chen, C.~Li, and R.-V. S{\'a}nchez, ``Multi-layer neural network with deep
  belief network for gearbox fault diagnosis.'' \emph{Journal of
  Vibroengineering}, vol.~17, no.~5, 2015.

\bibitem{gan2016construction}
M.~Gan, C.~Wang \emph{et~al.}, ``Construction of hierarchical diagnosis network
  based on deep learning and its application in the fault pattern recognition
  of rolling element bearings,'' \emph{Mechanical Systems and Signal
  Processing}, vol.~72, pp. 92--104, 2016.

\bibitem{oh2016dbn}
H.~Oh, B.~C. Jeon, J.~H. Jung, and B.~D. Youn, ``Smart diagnosis of journal
  bearing rotor systems: Unsupervised feature extraction scheme by deep
  learning,'' 2016.

\bibitem{zhang2015deep}
C.~Zhang, J.~H. Sun, and K.~C. Tan, ``Deep belief networks ensemble with
  multi-objective optimization for failure diagnosis,'' in \emph{Systems, Man,
  and Cybernetics (SMC), 2015 IEEE International Conference on}.\hskip 1em plus
  0.5em minus 0.4em\relax IEEE, 2015, pp. 32--37.

\bibitem{zhang2016multiobjective}
C.~Zhang, P.~Lim, A.~Qin, and K.~C. Tan, ``Multiobjective deep belief networks
  ensemble for remaining useful life estimation in prognostics,'' \emph{IEEE
  Transactions on Neural Networks and Learning Systems}, 2016.

\bibitem{janssens2016convolutional}
O.~Janssens, V.~Slavkovikj, B.~Vervisch, K.~Stockman, M.~Loccufier,
  S.~Verstockt, R.~Van~de Walle, and S.~Van~Hoecke, ``Convolutional neural
  network based fault detection for rotating machinery,'' \emph{Journal of
  Sound and Vibration}, 2016.

\bibitem{babu2016deep}
G.~S. Babu, P.~Zhao, and X.-L. Li, ``Deep convolutional neural network based
  regression approach for estimation of remaining useful life,'' in
  \emph{International Conference on Database Systems for Advanced
  Applications}.\hskip 1em plus 0.5em minus 0.4em\relax Springer, 2016, pp.
  214--228.

\bibitem{he2017cnn}
X.~Ding and Q.~He, ``Energy-fluctuated multiscale feature learning with deep
  convnet for intelligent spindle bearing fault diagnosis,'' \emph{IEEE
  Transactions on IM}, 2017.

\bibitem{guo2016hierarchical}
X.~Guo, L.~Chen, and C.~Shen, ``Hierarchical adaptive deep convolution neural
  network and its application to bearing fault diagnosis,'' \emph{Measurement},
  vol.~93, pp. 490--502, 2016.

\bibitem{lecun1998gradient}
Y.~LeCun, L.~Bottou, Y.~Bengio, and P.~Haffner, ``Gradient-based learning
  applied to document recognition,'' \emph{Proceedings of the IEEE}, vol.~86,
  no.~11, pp. 2278--2324, 1998.

\bibitem{wang2016}
J.~Wang, j.~Zhuang, L.~Duan, and W.~Cheng, ``A multi-scale convolution neural
  network for featureless fault diagnosis,'' in \emph{Proceedings of 2016
  International Symposium of Flexible Automation (ISFA)}, 2016, pp. 1--6.

\bibitem{chen2015gearbox}
Z.~Chen, C.~Li, and R.-V. Sanchez, ``Gearbox fault identification and
  classification with convolutional neural networks,'' \emph{Shock and
  Vibration}, vol. 2015, 2015.

\bibitem{weimer2016design}
D.~Weimer, B.~Scholz-Reiter, and M.~Shpitalni, ``Design of deep convolutional
  neural network architectures for automated feature extraction in industrial
  inspection,'' \emph{CIRP Annals-Manufacturing Technology}, 2016.

\bibitem{dongsmall}
H.-Y. DONG, L.-X. YANG, and H.-W. LI, ``Small fault diagnosis of front-end
  speed controlled wind generator based on deep learning.''

\bibitem{ince2016real}
T.~Ince, S.~Kiranyaz, L.~Eren, M.~Askar, and M.~Gabbouj, ``Real-time motor
  fault detection by 1-d convolutional neural networks,'' \emph{IEEE
  Transactions on Industrial Electronics}, vol.~63, no.~11, pp. 7067--7075,
  2016.

\bibitem{abdeljaber2017real}
O.~Abdeljaber, O.~Avci, S.~Kiranyaz, M.~Gabbouj, and D.~J. Inman, ``Real-time
  vibration-based structural damage detection using one-dimensional
  convolutional neural networks,'' \emph{Journal of Sound and Vibration}, vol.
  388, pp. 154--170, 2017.

\bibitem{7748035}
M.~Yuan, Y.~Wu, and L.~Lin, ``Fault diagnosis and remaining useful life
  estimation of aero engine using lstm neural network,'' in \emph{2016 IEEE
  International Conference on Aircraft Utility Systems (AUS)}, Oct 2016, pp.
  135--140.

\bibitem{lstmrui}
R.~Zhao, J.~Wang, R.~Yan, and K.~Mao, ``Machine helath monitoring with lstm
  networks,'' in \emph{2016 10th International Conference on Sensing Technology
  (ICST)}.\hskip 1em plus 0.5em minus 0.4em\relax IEEE, 2016, pp. 1--6.

\bibitem{ruiconvlstm}
R.~Zhao, R.~Yan, J.~Wang, and K.~Mao, ``Learning to monitor machine health with
  convolutional bi-directional lstm networks,'' \emph{Sensors}, 2016.

\bibitem{malhotra2016multi}
P.~Malhotra, A.~Ramakrishnan, G.~Anand, L.~Vig, P.~Agarwal, and G.~Shroff,
  ``Multi-sensor prognostics using an unsupervised health index based on lstm
  encoder-decoder,'' \emph{arXiv preprint arXiv:1608.06154}, 2016.

\bibitem{he2015deep}
K.~He, X.~Zhang, S.~Ren, and J.~Sun, ``Deep residual learning for image
  recognition,'' \emph{arXiv preprint arXiv:1512.03385}, 2015.

\bibitem{imagenet_cvpr09}
J.~Deng, W.~Dong, R.~Socher, L.-J. Li, K.~Li, and L.~Fei-Fei, ``{ImageNet: A
  Large-Scale Hierarchical Image Database},'' in \emph{CVPR09}, 2009.

\bibitem{maaten2008visualizing}
L.~v.~d. Maaten and G.~Hinton, ``Visualizing data using t-sne,'' \emph{Journal
  of Machine Learning Research}, vol.~9, no. Nov, pp. 2579--2605, 2008.

\bibitem{yosinski2015understanding}
J.~Yosinski, J.~Clune, A.~Nguyen, T.~Fuchs, and H.~Lipson, ``Understanding
  neural networks through deep visualization,'' \emph{arXiv preprint
  arXiv:1506.06579}, 2015.

\bibitem{pan2010survey}
S.~J. Pan and Q.~Yang, ``A survey on transfer learning,'' \emph{IEEE
  Transactions on knowledge and data engineering}, vol.~22, no.~10, pp.
  1345--1359, 2010.

\bibitem{shen2015bearing}
F.~Shen, C.~Chen, R.~Yan, and R.~X. Gao, ``Bearing fault diagnosis based on svd
  feature extraction and transfer learning classification,'' in
  \emph{Prognostics and System Health Management Conference (PHM), 2015}.\hskip
  1em plus 0.5em minus 0.4em\relax IEEE, 2015, pp. 1--6.

\bibitem{7542845}
J.~Xie, L.~Zhang, L.~Duan, and J.~Wang, ``On cross-domain feature fusion in
  gearbox fault diagnosis under various operating conditions based on transfer
  component analysis,'' in \emph{2016 IEEE International Conference on
  Prognostics and Health Management (ICPHM)}, June 2016, pp. 1--6.

\bibitem{lu2016deep}
W.~Lu, B.~Liang, Y.~Cheng, D.~Meng, J.~Yang, and T.~Zhang, ``Deep model based
  domain adaptation for fault diagnosis,'' \emph{IEEE Transactions on
  Industrial Electronics}, 2016.

\bibitem{mao2017online}
W.~Mao, L.~He, Y.~Yan, and J.~Wang, ``Online sequential prediction of bearings
  imbalanced fault diagnosis by extreme learning machine,'' \emph{Mechanical
  Systems and Signal Processing}, vol.~83, pp. 450--473, 2017.

\bibitem{duan2016new}
L.~Duan, M.~Xie, T.~Bai, and J.~Wang, ``A new support vector data description
  method for machinery fault diagnosis with unbalanced datasets,'' \emph{Expert
  Systems with Applications}, vol.~64, pp. 239--246, 2016.

\bibitem{huang2016learning}
C.~Huang, Y.~Li, C.~Change~Loy, and X.~Tang, ``Learning deep representation for
  imbalanced classification,'' in \emph{Proceedings of the IEEE Conference on
  Computer Vision and Pattern Recognition}, 2016, pp. 5375--5384.

\bibitem{yan2015deep}
Y.~Yan, M.~Chen, M.-L. Shyu, and S.-C. Chen, ``Deep learning for imbalanced
  multimedia data classification,'' in \emph{2015 IEEE International Symposium
  on Multimedia (ISM)}.\hskip 1em plus 0.5em minus 0.4em\relax IEEE, 2015, pp.
  483--488.

\end{thebibliography}

\end{document}